\documentclass[10pt,twocolumn,letterpaper]{article}

\usepackage{cvpr}      % To produce the REVIEW version

\usepackage{algorithm}
\usepackage{algorithmic}
\usepackage[table]{xcolor}
\usepackage{tcolorbox}
\usepackage{caption}
\usepackage{setspace}
\usepackage{courier}
\definecolor{codebg}{HTML}{E6F0E6}
\definecolor{codeblue}{HTML}{D9EAF7}
\usepackage{multirow}
\usepackage{stfloats}
\usepackage{booktabs}
\usepackage{xcolor}

\usepackage{makecell}
\definecolor{cvprblue}{rgb}{0.21,0.49,0.74}
\usepackage[pagebackref,breaklinks,colorlinks,allcolors=cvprblue]{hyperref}

\def\paperID{350} % *** Enter the Paper ID here
\def\confName{CVPR}
\def\confYear{2026}

\title{InstanceV: Instance-Level Video Generation}

\author{
Yuheng Chen$^1$
\quad Teng Hu$^1$
\quad Jiangning Zhang$^2$
\quad Zhucun Xue$^2$
\quad Ran Yi$^1$\\
\quad Lizhuang Ma$^1$\textsuperscript{\dag}\\
\normalsize $^1$Shanghai Jiao Tong University \quad $^2$Zhejiang University\\
{\tt\small Project page: \href{https://aliothchen.github.io/projects/InstanceV}{\textcolor{magenta}{https://aliothchen.github.io/projects/InstanceV}}}
\\
}

\begin{document}

\twocolumn[{%
\maketitle
\begin{figure}[H]
\vspace{-0.4in}
\hsize=\textwidth 
\centering
\includegraphics[width=1.0\textwidth]{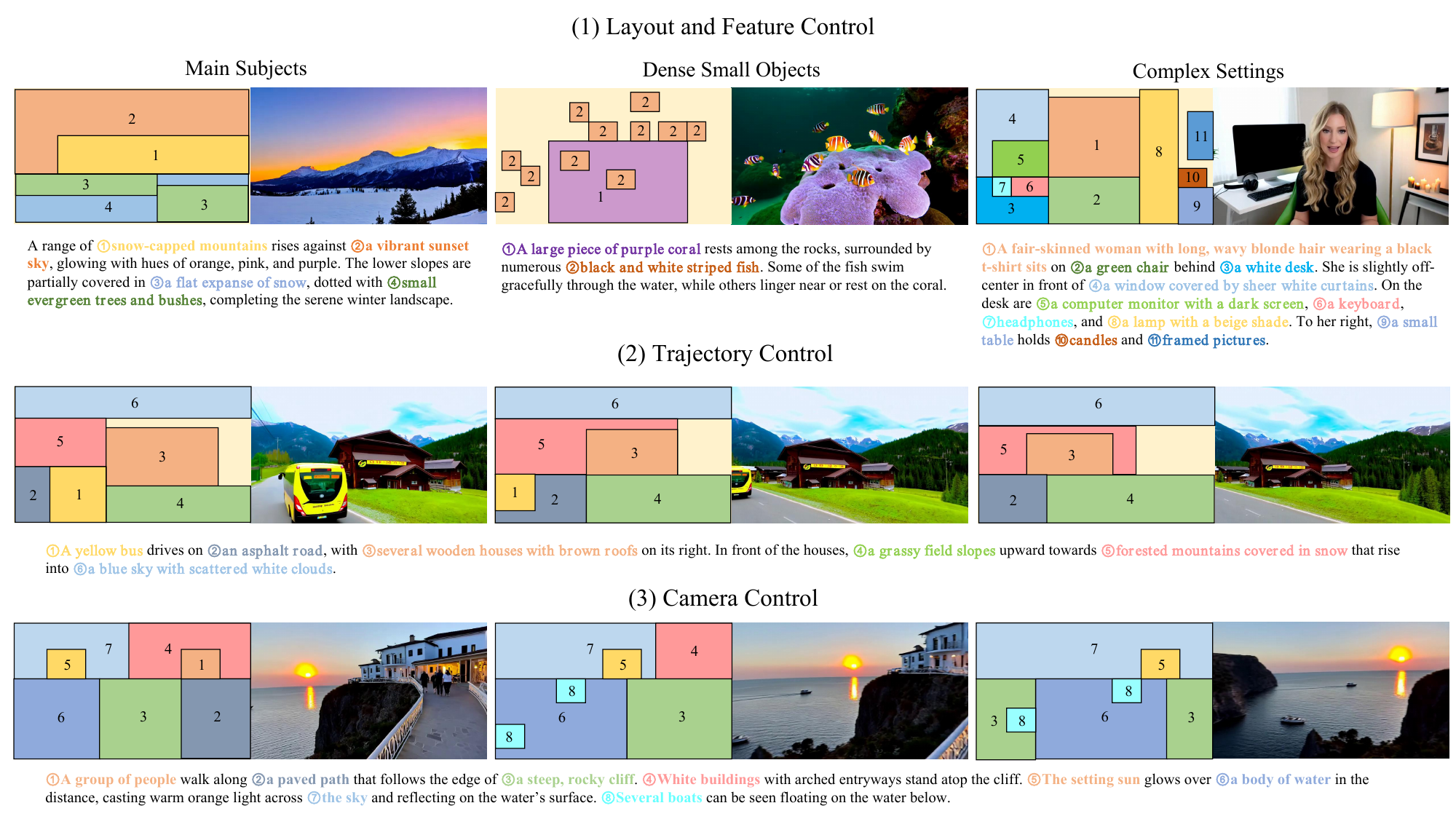}
\caption{
% instance-level grounding information不仅可以帮助video diffusion model更好的理解global video caption中layout，instance attribute等细粒度控制信息，还可以间接实现instance-level的trajectory control和global层面的相机控制。
The instance-level grounding information not only helps the video diffusion model better understand fine-grained control cues in the global video caption, such as instances layout and attributes, but also enables indirect realization of instance-level trajectory control and global camera motion control.
}
\label{fig:fig0}
% \vspace{-10pt}
\end{figure}
}]

\maketitle

{%
\renewcommand{\thefootnote}{}%
\footnotetext{\textsuperscript{\dag} Corresponding author.}%
}

% instwan改成instVDM?

\begin{abstract}
Recent advances in text-to-video diffusion models have enabled the generation of high-quality videos conditioned on textual descriptions. 
However, most existing text-to-video models rely solely on textual conditions,
lacking general fine-grained controllability over video generation.
To address this challenge,
we propose \textbf{InstanceV}, a video generation framework that enables \textbf{i)} instance-level control and \textbf{ii)} global semantic consistency. 
Specifically, with the aid of proposed Instance-aware Masked Cross-Attention mechanism, InstanceV maximizes the utilization of additional instance-level grounding information to generate correctly attributed instances at designated spatial locations.
To improve overall consistency, We introduce the Shared Timestep-Adaptive Prompt Enhancement module, which connects local instances with global semantics in a parameter-efficient manner.  Furthermore, we incorporate Spatially-Aware Unconditional Guidance during both training and inference to alleviate the disappearance of small instances. Finally, we propose a new benchmark, named InstanceBench, which combines general video quality metrics with instance-aware metrics for more comprehensive evaluation on instance-level video generation.
Extensive experiments demonstrate that InstanceV not only achieves remarkable instance-level controllability in video generation, but also outperforms existing state-of-the-art models in both general quality and instance-aware metrics across qualitative and quantitative evaluations.
% on both general-purpose video quality metrics and instance-aware evaluation metrics.
\end{abstract}    
% \begin{figure*}[!t]
%     \centering
% \includegraphics[width=\textwidth]{figures/fig0.pdf}
%     \caption{
%     % 我们提出的InstWan框架的instance-level控制能力的展示。（1）instance-level control最常见的是对instance的feature和视频整体的layout进行控制，分为main subjects layout、dense small objects 和complex settings等三个方面 （2）instwan同样可以对运动的instance实现精细的trajectory的控制 （3）instwan可以有效的实现不同相机movements下对instances的控制。
%     Demonstration of the instance-level control capabilities of our proposed InstWan framework.
% (1) The most common type of instance-level control involves manipulating both the features of individual instances and the overall video layout, which can be categorized into three aspects: main subjects layout, dense small objects, and complex settings.
% (2) InstWan also enables fine-grained trajectory control for moving instances, accurately guiding their motion patterns across frames.
% (3) In addition, InstWan effectively maintains instance control under various camera movements, demonstrating strong robustness to viewpoint and motion changes.
%     }
%     \label{fig:enter-label}
% \end{figure*}

\section{Introduction}
\label{sec:intro}

Driven by rapid advances in foundational backbones~\cite{peebles2023scalable} and multimodal interaction, a new generation of open-source~\cite{wan2025wan,kong2024hunyuanvideo,yang2024cogvideox,chen2025skyreels} 
has been introduced. These models not only substantially improve the visual quality, text-to-video cross-modal alignment and diversity of generated videos, but also demonstrate promising applications in real-world scenarios like education, design, and entertainment.
% giving rise to numerous popular online IPs and fostering diverse paradigms of human–computer interaction.

% 当prompt较为复杂，比如涉及到多个主体，主体之间存在交互和较为复杂的物理空间关系时，生成的视频可能无法准确地follow video caption，遗漏instance或者生成错误属性的instance，更严重地会导致生成结果存在一定的artifacts
% However, existing video generation frameworks share a notable limitation: 
% when the prompt becomes more complex, for example involving multiple subjects with interactions and intricate spatial relationships, the generated video may fail to align with the given prompt. This often results in missing or misattributed instances, and more severely, introduces noticeable artifacts.
However, existing video generation frameworks share a notable limitation: they struggle to maintain fidelity to complex prompts, particularly those specifying multiple interacting instances or intricate spatial relationships.
% due to their limitation in grounding informatino only support text. 
This limitation often manifests as semantic errors, such as missing instances or misattributed properties, and may result in noticeable visual artifacts in the generated videos.
Prior to the recent focus on video generation, extensive research in text-to-image (T2I) synthesis explored methods for fine-grained instance-level control
% ~\cite{zhou2024migc,sella2025instancegen,zheng2023layoutdiffusion,li2023gligen,wang2024instancediffusion}
% . However, adapting these T2I approaches to video is challenging. They 
, which can be broadly divided into three types, 
% each with significant limitations for video synthesis:
\textbf{(1)} \textit{\textbf{Implicitly conditioned methods}}~\cite{li2023gligen,wang2024instancediffusion,zheng2023layoutdiffusion} 
treat instance-level grounding information as additional visual tokens, which are injected through a residual self-attention mechanism.
Such methods require resource-intensive training, making them computationally prohibitive for most open-source video diffusion models.
% rely on resource-intensive training, requiring massive datasets and large batch sizes. This makes them computationally prohibitive for most video generation efforts, particularly within the open-source community.
\textbf{(2)} \textit{\textbf{First-instances-then-merge}}~\cite{zhou2024migc,wang2024instancediffusion} methods generate each instance separately before merging them together. These methods often require more denoising steps, and temporal consistencies are challenging to preserve in such manner.
\textbf{(3)} \textit{\textbf{Training-free}}~\cite{sella2025instancegen} approaches directly manipulate attention maps during inference. These methods pose significant challenges to current GPU hardware, as larger video feature maps result in substantially larger attention maps.
To address these challenges 
% of achieving instance-level control in video generation without incurring prohibitive computational costs
, we propose ~\textbf{InstanceV}, a computationally efficient and training-efficient video generation framework.
Specifically, we propose an ~\textbf{Instance-aware Masked Cross-Attention} (\textbf{IMCA}) module that seamlessly integrates instance-level grounding information into the backbone, serving as an explicit guidance signal for the native cross-attention. IMCA enables the synthesis of objects with specified attributes at desired spatial locations without additional denoising steps, while remaining computationally lightweight and easy to train.
To maintain global semantic consistency and temporal coherence, we further design a ~\textbf{Shared Timestep-Adaptive Prompt Enhancement} (\textbf{STAPE}) module, which effectively bridges local instance semantics with the global video caption in a parameter-efficient manner.
Furthermore, we incorporate a ~\textbf{Spatially-Aware Unconditional Guidance} (\textbf{SAUG}) strategy in both training and inference. By leaving the instance prompts empty, this mechanism encourages the model to better reason about spatial relationships among instances, thereby refining the generation of smaller instances. Finally, to better evaluate instance-level video generation, we propose a comprehensive benchmark named InstanceBench, which consists of both general video quality  and instance-level evaluation metrics for a more holistic assessment.
% which is able to 在指定的location生成指定属性的instance，与整体的video caption保持一致的语义，保持视频帧之间的一致性，with a single forward pass。
% that focuses on fine-grained instance-level control while preserving global consistency. 
% IMCA: Instance-aware Masked Cross Attention ，
% 为了在指定的位置能够生成正确属性的instances，InstanceV包含了一个
% we first introduce a Masked Instance Visual Cross-Attention mechanism, which employs masked cross-attention to enable precise and explicit control signal injection. To reduce training cost, the cross-attention layers are initialized with the pretrained weights from the native DiT backbone.
% (2) we then propose an Enhanced Instance–Prompt Module, which promotes better coordination between the newly injected instance information and the original text embeddings, thereby mitigating potential artifacts caused by representational “conflicts”.
% (3) to address the issue of small-instance disappearance arising from spatial overlap with large instances, we design a float-version attention mask along with an instance prompt classifier free guidance training mechanism.
% (4) finally, to improve both training efficiency and performance, we develop a data preparation pipeline that annotates not only instance-level prompts but also global contextual information, ensuring temporal consistency in grounding signals. In addition, we propose a preprocessing strategy that enhances user interaction and overall usability.

% 通过进行全面的实验，将我们的instanceV框架与wan-series model结合，是第一个模型层面的通用instance-level细粒度视频生成控制框架。
We conduct extensive experiments by extending the Wan-series~\cite{wan2025wan} models with our proposed InstanceV framework. Comparison against leading state-of-the-art models, including Wan-series, HunyuanVideo~\cite{kong2024hunyuanvideo}, CogVideoX~\cite{yang2024cogvideox}, and SkyReelsV2~\cite{chen2025skyreels}, demonstrate that \textit{InstanceV} surpasses all competing models on InstanceBench, showing superior performance on both general video quality metrics and specialized instance-aware metrics. Our main contributions are fourfold:
% which is the first architecture-level approach to support instance-level control in video generation
\begin{itemize}     
    \item We propose InstanceV, a computationally efficient yet powerful video generation framework and, to the best of our knowledge, the first to be designed specifically for instance-level control at the architectural level.
    \item We introduce Instance-aware Masked Cross-Attention, which explicitly associates instance-level prompts with their corresponding visual tokens, thereby enabling the precise and reliable generation of correct spatial layouts and richly detailed instance-specific attributes.
    \item We present a Shared Timestep-Adaptive Prompt Enhancement module to effectively mitigate potential semantic inconsistencies. Furthermore, we substantially enhance small-instance generation by incorporating a Spatially-Aware Unconditional Guidance strategy.
    \item To facilitate rigorous and comprehensive evaluation, we introduce InstanceBench, a new benchmark specifically designed for instance-level video generation. 
\end{itemize}

\section{Related Works}
\label{sec:relatedworks}

% 参考 polyvivid ultragen wan
\subsection{Video Diffusion Foundation Models}
% Driven by advances in generative modeling, the landscape of large-scale video models has evolved significantly, particularly in diffusion-based frameworks.
% 特别地，当DiT framework被提出之后，迅速成为主流的diffusion网络架构，同时DiT优异的可扩展性也激励开源社区朝着这一方向发展。近期，Wan series和HunyuanVideo通过扩展DiT架构，并在大规模的高质量文本-视频数据上进行训练，取得了impressive的视频生成效果。尽管如此，这些视频生成模型的grounding输入过于单一，只依赖文本的输入往往无法生成真正符合用户期望的视频结果。
% Driven by the rapid advances in generative modeling, the landscape of large-scale video generation has evolved substantially, particularly within diffusion-based frameworks.
% In particular, after the introduction of the DiT framework~\cite{peebles2023scalable}, it rapidly became the mainstream architecture for diffusion-based generation models. The remarkable scalability of DiT has further inspired the open-source community to advance in this direction
% Recently, models such as the Wan series~\cite{wan2025wan} and HunyuanVideo~\cite{kong2024hunyuanvideo} have extended the DiT architecture and achieved impressive video generation performance through large-scale training on high-quality text–video datasets.
% Nevertheless, the grounding inputs of these models remain overly simplistic, as they are solely conditioned on textual descriptions, which often prevents them from producing videos that truly align with the user’s expectations.

Driven by rapid advances in generative modeling, diffusion-based frameworks~\cite{opensora,hu2025ultragen,hu2025hunyuancustom,huang2025videomage} have become the widely adopted architecture for video generation, among which DiT ~\cite{peebles2023scalable} plays a central role owing to its outstanding scalability. 
Recent models, such as the Wan series~\cite{wan2025wan} and HunyuanVideo~\cite{kong2024hunyuanvideo}, further extend DiT and achieve impressive video generation performance through large-scale training on high-quality text–video datasets. 
Nevertheless, these models struggle when confronted with complex video captions depicting multiple interacting instances and intricate spatial relationships.

\subsection{Controllable Video Diffusion}
% 有一些现有的工作从不同角度扩充diffusion model的grounding信息。比如camera，motion，trajectory等通用条件控制，以及subjects-和ID-preserving等针对人或者其它主体的控制。
% 方法总结和归类
% pose: EchoMimicV2(half-body control，pose and audio),MagicPose(id and pose, pose net, appearence net, pretraining, multi-source self-attn),
% camera: Diffusion as Shader(DaS) (unify control, 3D video input, view 2d as projection of 3d), VD3D(camera-control for DiT)
% audio: Hallo2(hour-long, talking head),FantasyTalking（talkinghead，face lmk.body joints,audio encoder,face encoder）
% universal: VACE(three tasks, VEU)
% trajectory: Tora(traj-extractor, attention between blcoks),MotionCanvas(I2V, camera movements, scene-space object motions, with costly training data)
% ID: polyvivid(multi subject, rope, id-preserving), Identity-Preserving Text-to-Video Generation by Frequency Decomposition(face parsing, Q-former,)
% PolyVivid
Several existing works have sought to extend the grounding information of video diffusion from diverse perspectives, including general conditional controls such as camera~\cite{gu2025diffusion,bahmani2024vd3d}, motion~\cite{meng2025echomimicv2,chang2023magicpose}, audio~\cite{cui2024hallo2,wang2025fantasytalking}, and trajectory~\cite{zhang2025tora,xing2025motioncanvas}, as well as subject- and id-preserving controls~\cite{hu2025polyvivid,yuan2025identitypreserving} that focus on humans or other specific entities. 
% 这些工作都需要获取特定的控制条件，并通过增加的模块提取控制信息并注入到backbone中。因此，部分工作也面临着对数据依赖性高，缺乏大规模高质量的特定训练数据，以及增加的模块结构过于复杂难以训练的挑战，并且控制生成的内容在某种程度上有很大的局限性。值得一提，有一些工作通过统一的框架实现了多种控制条件的注入。VACE通过额外的Video Condition Unit实现了reference-to-video generation, video-to-video editing, and masked video-to-video editing多种视频生成控制任务的统一。DaS通过额外的3D tracking videos实现了统一生成控制，提出2d视频是3d内容的一个投影。
These approaches typically require the acquisition of specific control conditions, which are subsequently injected into the diffusion backbone via additional modules. 
% Consequently, many of them suffer from strong data dependency, a lack of large-scale, high-quality, task-specific training data~\cite{gu2025diffusion,xing2025motioncanvas,wang2025fantasytalking}, and the increased structural complexity of the added specialized modules~\cite{chang2023magicpose,wang2025fantasytalking}, which makes them difficult to train effectively and generalize well. 
Consequently, many of them are heavily dependent on large-scale, high-quality, task-specific training data~\cite{gu2025diffusion,xing2025motioncanvas,wang2025fantasytalking}, and the increased structural complexity of the specialized modules~\cite{chang2023magicpose,wang2025fantasytalking} further limits effective training and generalization.
Furthermore, the generated content conditioned on certain grounding information remains highly constrained~\cite{meng2025echomimicv2}, whereas subject- and id-preserving approaches~\cite{yuan2025identitypreserving,hu2025polyvivid} primarily focus on controlling larger instances.
Notably, some works aim to unify multiple formats of grounding information within a single framework. For instance, VACE~\cite{Jiang2025VACE} introduces a Video Condition Unit (VCU) for task-agnostic control, while DaS~\cite{gu2025diffusion} leverages 3D-tracked videos for generation, treating 2D videos as projections of 3D content.
In contrast, InstanceV achieves a simple yet effective form of instance-level control according to Table.~\ref{tab:lightweight}, which also extends naturally to support a broader range of control types.

% 尽管这些工作在不同的角度都取得了惊艳的控制效果，但是据我们所知仍然没有在instance-level或者控制layout的视频生成控制方法。

\begin{table}[t]
    \centering
    \resizebox{0.95\linewidth}{!}{
    \begin{tabular}{c|ccc}
    \toprule
        \textbf{Method} & \textbf{FLOPS} $\downarrow$ & \textbf{Params} $\downarrow$ & \textbf{Time Cost} $\downarrow$ \\
         \midrule
GLIGEN~\cite{li2023gligen} & 102.79\% &26.13\%& 96.83\%\\
    InstanceDiffusion~\cite{wang2024instancediffusion}     & 121.82\%&42.94\%& 115.69\%\\
    InstanceGen~\cite{sella2025instancegen} & 40.00\% & \textemdash & 80.00\% \\
    \textbf{InstanceV(ours)} &\textbf{15.02\%}&\textbf{20.65\%}&\textbf{9.14\%} \\
    \bottomrule
    \end{tabular}
    }
    \caption{Additional FLOPs, parameters, and time overhead (\%) introduced for instance-level control relative to the backbone.}
    \label{tab:lightweight}
    \vspace{-10pt}
\end{table}

\subsection{Instance-Level Control}
% 在T2I任务中，存在着一些instance-level和控制方法。CLIGEN是这一系列工作的先驱者，通过新增self-attention分支注入instance semantics和与之对应的spatial configuration。InstanceDiffusion在CLIGEN的基础上，丰富了注入的spatial configuration的形式。MIGC独立每个instance的生成，将instance results合并后得到最终的生成结果。
% 此外，也有一些training-free的生成框架，比如instance-cap通过构造structured caption，instancegen通过对前向过程中的attention map进行reweight和masking来实现控制效果。

For text-to-image generation, several instance-level control methods have been proposed. CLIGEN~\cite{li2023gligen} serves as an early pioneer in this line of work, which fuses instance features and spatial locations via neural networks and injects them into the backbone through a residual self-attention module. InstanceDiffusion~\cite{wang2024instancediffusion} extends CLIGEN by enriching the forms of injected spatial configurations. MIGC~\cite{zhou2024migc} features generating each instance independently followed by subsequent fusions. LayoutDiffusion~\cite{zheng2023layoutdiffusion} maps visual tokens and instance information into a shared latent space to facilitate more effective interaction. In addition, InstanceGen~\cite{sella2025instancegen} manipulates the generation process by reweighting and masking attention maps during inference.
For text-to-video tasks, InstanceCap~\cite{fan2025instancecap} represents the first attempt from a data-driven perspective, introducing an instance-level dataset annotated by a proposed instance-aware captioner. Compared to above works, InstanceV is more lightweight and computationally efficient, achieving more direct instance-level control benefited from a specialized model design.
\section{Instance-Level Data Preparation}
\label{sec:data_preparation}
% allowing for more instance-level information injection，我们需要构建一个数据预处理pipeline来从已有的text-video pairs中获取尽可能准确的instance信息。
To allow for richer instance-level information injection, we design a data preprocessing pipeline that extracts the most accurate instance information possible from existing text–video pairs.
% \subsection{Foreground}
\subsection{MLLM-based Instance Partitioning}
% 为了获取foreground中不同instance的语义信息和位置信息，我们使用text-video pairs来guide MLLM进行语义分割。具体地，我们应用了Florence2模型，pipeline如下 1）我们使用florence2的phrase_grounding_segmentation任务来初步划分出不同instance的语义信息和位置信息 
% 2）由于LLM(MLLM)对数量信息并不敏感，florence2倾向于将位置相近的同类个体划分为一个整体,类似crowd和a group of的语义，这有害于模型的训练。此外，在比较复杂的场景中，florence2模型会出现遗漏instance的情况。因此我们融合了florence2的不同tasks的结果。
% 2) 当视频中出现多个相同语义的objects时，对应的text caption通常会使用a crowd of或者a group of，或者仅仅是复数名词等类似的语义来进行表示，即使video中对应objects的数量并不多，比如3~5个，那么florence2会将这些objects统一归为一个instance。其中bounding box就代表这些物体整体的位置，涵盖的tokens数量可能远多于实际情况，会造成位置信息误差较大。并且对应的instance prompts的语义信息在数量上也比较模糊，造成LLM理解困难。
\noindent\textbf{Foreground.}
To obtain both semantic and spatial information of different foreground instances, 
% we employ text–video pairs to guide a multimodal large language model (MLLM) for semantic segmentation. Specifically, 
we adopt the \textit{Florence-2}~\cite{xiao2024florence2} together with \textit{Grounded-Sam2}~\cite{ren2024grounded,ravi2024sam}, and the pipeline proceeds as follows:
\textbf{(1)} We first utilize the phrase grounding segmentation task to initially identify the textual prompts and spatial locations of individual instances.
% (2) Since large language models (LLMs) and multimodal large language models (MLLMs) are generally insensitive to numerical distinctions, Florence-2 tends to merge nearby objects of the same category into a single entity, producing semantics similar to “a crowd” or “a group of.” Such behavior is detrimental to model training, as it obscures fine-grained instance information. 
\textbf{(2)} We then integrate the outputs from more tasks, such as object detection, to obtain more accurate and comprehensive instance segmentation results. 
% When multiple objects with the same semantics appear in a video, the corresponding text captions usually describe them using phrases such as “a crowd of”, “a group of”, or simply plural nouns. However, even when the actual number of objects is small (e.g., 3–5), Florence-2 tends to merge them into a single instance. In this case, the bounding box represents the overall region covering all these objects, which may affects many visual tokens not belonging to this instance, thus introduces significant spatial inaccuracies. Also the semantic information encoded in the corresponding instance prompts becomes vague in terms of quantity, making it more difficult for the LLM to correctly interpret the scene.
% Moreover, in complex scenes, Florence-2 may miss certain instances entirely. 
Two limitations remain in the previous results:
1) when multiple objects with similar semantics appear (e.g., “a crowd of people”), Florence-2 tends to merge them into a single large instance, resulting in ambiguous semantic representations; and
2) large instances that are partially occluded by other objects may be completely missed.
By integrating the complementary outputs from different tasks, more accurate and complete instance-level grounding information can be obtained.
% To mitigate these issues, we integrate the outputs from multiple Florence-2 tasks to obtain more comprehensive instance segmentation results.
% 3）与图片不同，视频中会出现部分instances出现或者消失的情况。所以当逐帧对video进行instance-level的语义分割时，florence-2 无法精准把握时序语义，tends to hungrily attach an object to a most-similar phrase in global caption，尽管可能在那一帧中phrase表示的intance并未出现或者已经消失。因此我们使用clip-score来进一步prun掉潜在的错误instance segmentation。
% \textbf{(3)} We further employ a CLIP-score–based~\cite{radfordLearningTransferableVisual2021a} filtering strategy to prune potentially erroneous instance segmentations. Unlike images, videos inherently involve the appearance and disappearance of certain instances over time. Consequently, when performing frame-by-frame instance-level semantic segmentation, Florence-2 often struggles to accurately capture temporal semantics. It tends to greedily associate a phrase in the global caption with a similar but wrong object in the current frame, even when the instance referred to by that phrase may not exist, or may have already disappeared. 
\textbf{(3)}
We further employ a CLIPScore–based~\cite{radfordLearningTransferableVisual2021a} filtering strategy to prune potentially erroneous instance segmentations.
Unlike static images, videos naturally exhibit temporal variations, where certain instances may newly appear or disappear across frames.
Using the same text description to process all frames can therefore lead to mislabeling.
The CLIPScore-based filtering helps mitigate such problems by removing mismatched instances.
% \subsection{Background}
% 之前的instance-level T2I工作中，background信息通常被忽略了，但是我们认为background可以被看做是一个巨大的instance。并且，background-instance可以不仅仅只包括背景中所包含的实体，也可以是任何更加广泛的global的概念，包括相机参数、画面风格等。
% 我们通过instruct LLM来根据video caption总结得到background的语义信息。一个frame中不代表任何instance的区域被认为是background instance。

% In previous instance-level text-to-image (T2I) works, background information was often overlooked. However, we argue that the background can be regarded as a large-scale instance in itself. Moreover, the background instance is not limited to the physical entities present in the scene, but can also encompass broader global concepts, such as camera parameters, overall visual style, or other holistic scene attributes.
\noindent\textbf{Background.}
To facilitate a more straightforward and coherent implementation of InstanceV, we utilize a background prompt derived from the summarized video caption by an LLM~\cite{yang2024qwen25} as the instance information for regions not associated with any specific foreground instances. 
% The instance-level grounding information obtained at this stage consists of textual instance prompts and spatial location information in the form of bounding boxes.
% 关于二者如何处理为model可接受的输入，参考section methods and 补充材料
At this stage, the instance-level grounding information includes textual instance prompts and spatial coordinates in the form of bounding boxes.
Please refer to Section~\ref{sec:methods} and the supplementary materials for details on how these are converted into model inputs.

% 这一步得到的intance-level grounding information的形式是文本的instance prompts和bounding-box形式的位置信息。
% 扩充grounding information的信息来源，分为不同的难易等级，1) point(user-friendly) 2) bbox(general format) 3) segmaps (for demanding uses)
% \subsection{Format Expansion}
% 现在我们已经得到了instance的文本形式和语义信息和bounding boxes形式的位置信息。但是对于多样化的控制需求来说，instance information形式上有一些单一。因此，我们还准备了一定比例的point和segmentation mask的位置信息，以便让我们的模型能够支持简易位置控制和精确位置控制。对于point而言，我们instruct一个LLM来根据text caption和single point 坐标随机生成一个合理的bbox。for segmentation masks，我们用已有的bbox来ground sam2生成segmentaion mask。

% vae降维加上dit的token patchify，详细说明一下维度
\subsection{Downscale and Patchifying}
% 主流的DiT设计中都包含一个VAE模块，用于聚合相邻像素的信息，降低visual tokens的数量，加速训练和推理。对于Video diffusion backbone而言，除了在空间维度上降低复杂度，还需要在时间维度上聚合相邻帧的信息。假设VAE在temporal维度上的downscale为T_ds，那么data preprocessing阶段，假设0-indexed，我们只处理得到（0，T_ds，2*T_ds，...）帧的额外grounding information，因为我们认为在时间跨度范围较小的情况下，instaces的数量和位置不会出现非常剧烈的变化。这样的处理方式有助于减小data preprocessing的开销，降低model的结构复杂性，以及grounding information更加直接的注入。

% In mainstream DiT architectures, a VAE module is typically employed to aggregate information from neighboring pixels, thereby reducing the number of visual tokens and accelerating both training and inference.
% For video diffusion backbones, however, it is also necessary to aggregate information across adjacent frames in the temporal dimension, in addition to reducing spatial complexity.
In mainstream video diffusion models, videos are represented as discrete visual tokens obtained through a VAE and a patchifying module, where both spatial and temporal dimensions are compressed. Spatially, the bounding boxes can be directly rescaled.
Temporally, let $T_{ds}$ denote the temporal downscaling factor of the VAE. Assuming 0-indexing, we extract additional grounding information only from frames $(0, T_{ds}, 2T_{ds}, \ldots)$. This design relies on the assumption that within a short temporal stride, the number and spatial distribution of instances remain relatively stable. Such a strategy reduces the computational overhead of data preprocessing, simplifies the model complexity, and allows for more straightforward injection of instance-level grounding information.

% 
% \subsection{Prunning}
% 在开源text to video数据集中，存在着一定量的包含场景切换的视频，意味着存在相邻帧之间发生视觉内容的突变，包括亮度、颜色分布、纹理、语义等的大幅变化，这也意味着instance信息也会发挥剧烈变化。一个视频的不同片段有可能包含完全不同的instances，这对基于FLorence2等MLLM的预处理和后续的训练都造成了极大的挑战。因此我们在构建训练数据集时排除掉这部分数据（transnetv2）。同时，对于高帧率的视频，有可能我们选择的用于训练的片段中缺乏motion信息，呈现较为静态的视觉效果。我们对这一部分视频采用了降低帧率的策略，来保证我们构建的instance-level的训练数据集中，包含拥有丰富instance动态语义的视频。
% In open-source text-to-video datasets, a non-negligible portion of videos contain scene transitions, where adjacent frames exhibit abrupt visual changes in terms of brightness, color distribution, texture, and semantics. Such discontinuities also imply drastic variations in instance information, as different segments of a video may involve entirely distinct instances.
% This poses significant challenges for both MLLM-based preprocessing (e.g., using Florence2) and the subsequent training process.
% To mitigate this issue, we filter out such samples using TransNetV2 during dataset construction.
% Moreover, for high frame rate videos, the selected clips may lack sufficient motion cues, leading to overly static visual dynamics.
% To address this, we adopt a frame-rate reduction strategy, ensuring that our instance-level training dataset contains videos with rich instance dynamics and meaningful temporal variations.
\begin{figure*}[!t]
    \centering
    \includegraphics[width=\textwidth]{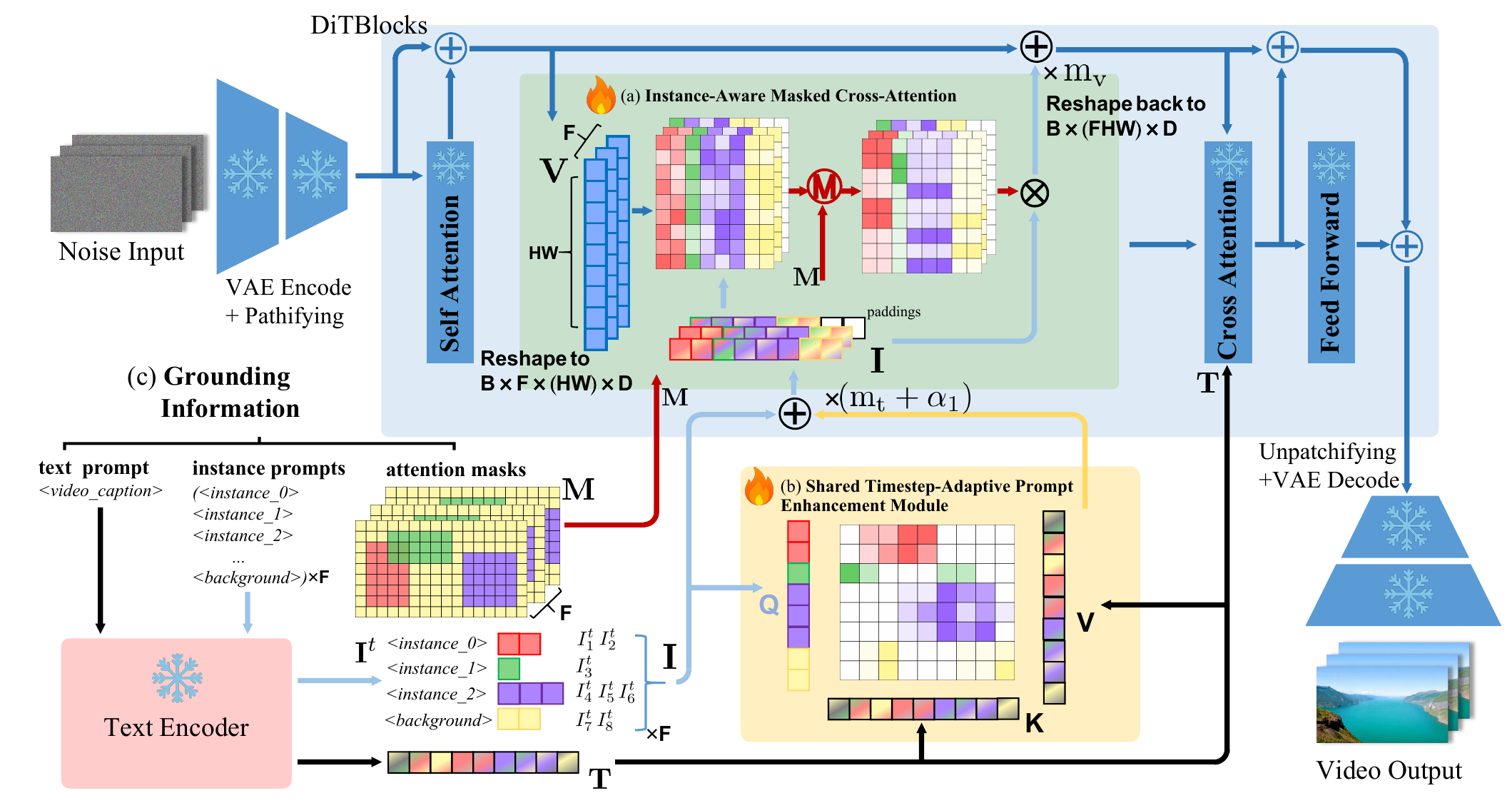}
    \caption{
    % 我们提出的InstWan框架的概览。所有的backbone模块在训练时都是frozen的。经过VAE encode和dit的patchifying操作后，得到了B*(FHW)*D的feature map。为了简化示意图我们省略了Dit中的text embeddings和所有的layernorm层。
    % （a）中由于空间限制我们只绘制了很小一部分visual tokens，可以将这里的visual tokens图示理解为H=1时，其中的attention maps可以体现出实际情况的attention maps。（b）中我们使用纯色来表示分开编码的instances prompts，而video caption中对应的tokens因为注意力的存在，实际编码中包含了更多的全局语义，我们使用渐变色来进行表示。
    % （c）中所示grounding information中attention masks和instance prompts也是F组。并且由于时序原因，每一组包含的instances prompts并不相同。需要特别注意的是，为了便于理解这里我们attention masks绘制为F*H*W，仅用颜色区分不同instances，但是实际实现中在不考虑multiple heads的情况下，attention masks的shape应该为F*HW*N_ins，其中N_ins为最大instance tokens序列长度，不足会进行paddings。由于我们不再使用clip进行instances特征提取，而是backbone中的text encoder，通常为一个LLM，所以不同instances对应的tokens数量并不相同。
    Overview of the proposed InstanceV framework.
    % All backbone modules are frozen during training. 
    % After VAE encoding and DiT patchifying, we obtain a feature map of size $B \times (FHW) \times D$. For clarity, the text embeddings layers and all LayerNorms in the DiT backbone are omitted from the illustration.
    \textbf{(a)} Only a small subset of visual tokens is shown due to space limitations; this visualization can be interpreted as a case where $H=1$. 
    % allowing the depicted attention maps to faithfully reflect the real spatial distribution. 
    \textbf{(b)} Solid colors denote independently encoded instance prompts, while the tokens with color encode more global semantic information.
    \textbf{(c)} The grounding information consists of $F$ groups of attention masks and instance prompts, where the instances differ across frames due to temporal variation. Note that for simplicity, the attention masks are drawn as $F \times H \times W$, with colors distinguishing different instances. 
    % In practice, ignoring the multiple attention heads, their actual shape is $F \times HW \times N_{\text{ins}}$, where $N_{\text{ins}}$ is the maximum length of instance tokens (padded as needed).
    % Unlike previous works that use CLIP for instance feature extraction, we reuse the text encoder from the backbone (typically an LLM), resulting in variable token lengths across different instances.
    }
    \label{fig:method}
    \vspace{-10pt}
\end{figure*}

\section{Methods}
\label{sec:methods}

To enable instance-level control, we make \textit{three key modifications} to the DiT backbone. First, by introducing the \textbf{Instance-aware Masked Cross-Attention (IMCA)}, instance-level grounding information is efficiently injected into the model, guiding the cross-attention module to generate correctly attributed instances at specified spatial positions. Second, to mitigate potential global semantic inconsistencies that may arise from IMCA, we propose a \textbf{Shared Timestep-Adaptive Prompt Enhancement (STAPE)} module, which connects local instances with global semantics and reduces artifact generation. Finally, to further improve the generation of small instances, we incorporate \textbf{Spatially-Aware Unconditional Guidance (SAUG)} during both training and inference, enhancing their fidelity and overall visual quality. For clarity, we omit all normalization layers and the batch dimension, and all indices are 1-indexed in the following description.

% DiT paper
% 在介绍方法前，我们首先定义一些符号。对于主流的DiT video diffusion backbones来说，一个DiT Block通常包含一个self-attn，一个cross-attn，和一个ffn。假设经过VAE encoding和token patchifying后，我们得到了NV个维度为D的tokens，NV=F*H*W，F为降维后的时间序列长度。经过text-encoder和DiT自身的text embedding后的prompt为NC*D_ctx。同时，经过data preparation后，我们得到了F组instance prompts，形状为(F,NI,D_Ins)，和F组attention masks，形状为(F,NI,NV)。

% After passing through the text encoder and the DiT’s own text embedding layer, the processed video caption is represented by a sequence of $N_{ctx}$ tokens, denoted as $\mathbf{T} = ({T_1, T_2, \ldots, T_{N_{ctx}}})$, where each token $T_i \in \mathbb{R}^{D_{ctx}}$. 
% The video caption is processed by a text encoder and the DiT's text embedding layer, yielding a sequence of $N_{ctx}$ tokens, denoted as $\mathbf{T} = ({T_1, T_2, \ldots, T_{N_{ctx}}})$, $T_i \in \mathbb{R}^{D_{ctx}}$.
% Meanwhile, after the data preparation stage discussed in Sec.\ref{sec:data_preparation}, we obtain $F$ groups of instance prompts tokens $\mathbf{I} = \{({I_1^1, I_2^1, \ldots, I^1_{N_{ins}}}), ({I_1^2, I_2^2, \ldots, I^2_{N_{ins}}}), \ldots, \}$, each group consisting of $N_{ins}$ tokens with a dimensionality of $D_{ins}$, and $F$ corresponding groups of attention masks $\mathbf{M}\in \mathbb{R}^{F\times N_{ins}\times N_{v}}$, whose detailed usage and definition will be further explained in Secs.~\ref{sec:MIV-CA} and~\ref{sec:FloatMask}.
% Additionally, we generate a corresponding attention mask , whose role and formulation are detailed in Secs.~\ref{sec:MIV-CA}.

\subsection{Instance-Aware Masked Cross-Attention}
% 在DiT backbone中，cross attention承担着不同模态交互以及为生成的视频赋予语义的责任.cross attention有着很强的自主性，其注入的语义信息往往很强烈。为了能够达到instance-level的细粒度控制效果，我们需要对cross attention施加控制，
In the DiT backbone, the Cross-Attention module is responsible for cross-modal interactions and determining the semantics of generated videos~\cite{hertz2022prompt,wen2025analysis,liu2024video}, thus achieving fine-grained instance-level control necessitates imposing guidance upon this powerful mechanism.
\label{sec:MIV-CA}
% 之前的用于Image-Level的Instance control的方法的instance特征一般选择使用clip的编码特征，并且会通过神经网络与位置信息进行融合。融合后的特征会作为额外的token拼接在visual tokens后，通过增加的self-attention模块以residual的形式添加到backbone的feature中。我们根据视频生成的diffusion backbone的特征对这种framework做出了改进。

\noindent\textbf{Explicit Instance Prompt Injection.}
% 在DiT框架中，输出视频的语义信息完全由cross-attention module决定。为了使得我们注入的instance-level information能够对cross-attn产生尽可能大的影响，最直接方式就是先cross-attn一步，为需要的visual token注入与cross-attn的domain 相同或者类似的语义信息。
% Previous methods for image-level instance control typically adopt CLIP-encoded features for instance representations, which are then fused with positional information through a neural network. The fused features are appended as additional tokens to the visual tokens and incorporated into the backbone’s features via an added self-attention module in a residual fashion. 
% 这存在几个问题，（1）feature leakage issues in CLIP embeddings~\cite{feng2022training}，这在longer video caption时会更加严重 （2）使用一个neural network来汇合position information和clip-encoded instance feature，并与visual token的语义进行对齐给训练增加了很大难度。
To maximize the influence of our instance-level information on cross-attention, the most efficient way is to inject the target visual tokens with semantic content that aligns with the context domain of following cross-attention module.
Supposing that  we obtain a set of $N_v$ visual tokens, $\mathbf{V} = ({V^1_1, V^1_2, \ldots, V^F_{HW}})$, $V^t_i \in \mathbb{R}^D$ after VAE encoding and DiT patchification.
% where each token $V^t_i \in \mathbb{R}^D$ is a $D$-dimensional feature vector. 
The total number of tokens is $N_v=F\times H\times W$, corresponding to a latent feature map with temporal length $F$ and spatial resolution $H \times W$. 
Previous methods for image-level instance control typically adopt CLIP-encoded features for instance representations, which are then fused with positional information through a neural network,
\begin{align}
        \overline{\mathbf{V}}&=[{\mathbf{V}}, \mathrm{MLP(Instance_{CLIP},Position)}] \label{eq:1-1}
\end{align}
The fused features are appended after visual tokens and incorporated into the backbone’s features via an added self-attention module in a residual fashion,
\begin{align}
\mathbf{V}&=\mathbf{V}+\mathrm{SelfAttn}(\overline{\mathbf{V}},\overline{\mathbf{V}})[:\mathrm{len(\mathbf{V})}] 
\end{align}
However, this approach suffers from two key limitations, (1) the feature leakage issues inherent in CLIP embeddings~\cite{feng2022training} will be exacerbated in tasks involving longer video captions; (2) tasking a single neural network to simultaneously fuse positional information with instance features and align the resulting representation with backbone visual semantics introduces significant training complexity.

% 为了充分地利用backbone的文本编码能力和跨模态语义对齐能力，我们使用backbone中的text-encoder和text embedding layer来编码分隔的instance prompts。同时不再将instance位置信息编码到intance prompt tokens中来以避免任何可能的对instance语义的干扰。
To fully leverage the backbone's text encoding and cross-modal semantic alignment capabilities, we employ its native text encoder and text embedding layer to process the individual instance prompts to get instance prompt tokens $~\mathbf{I} = \{\mathbf{I}^1, \mathbf{I}^2, \ldots, \mathbf{I}^F\}$. Each set $\mathbf{I}^t=(I_1^t, I_2^t, \ldots, I^t_{N_{ins}})$, $I_i^t \in \mathbb{R}^{D_{ins}}$, contains up to $N_{ins}$ tokens, with any shorter ones being zero-padded. Concurrently, we no longer encode positional information into $\mathbf{I}$, thereby avoiding any potential interference with their underlying semantics.

\noindent\textbf{Instance-Aware Cross-Attention.}
% since visual tokens $~\mathbf{V}$ 比 instance prompt tokens $~\mathbf{I}$要多得多，将I append在V的后面然后使用self attention机制进行信息交互对于instance-level的information injection是低效的，因为visual token之间显然有着更高的相似度。同时在video diffusion model中，self attention比cross attention要显著增加更多的FLOPS和time cost。
% 因此我们提出instance aware cross attention module，在减少计算量的同时把instance prompt tokens的语义最大限度地融合到visual tokens中去。instance aware cross attention以residual connection的形式添加到backbone中，受到一个0初始化的gated parameter $\mathbf{M_v}$的控制，以保证训练的稳定性。
Since visual tokens $\mathbf{V}$ significantly outnumber instance prompt tokens $\mathbf{I}$, appending $\mathbf{I}$ to $\mathbf{V}$ and using a self-attention mechanism for information exchange is inefficient for instance-level injection. This inefficiency arises because visual tokens exhibit high self-similarity, causing them to dominate the attention mechanism. Furthermore, in video diffusion models, self-attention incurs substantially higher computational costs (in both FLOPS and time cost) compared to cross-attention.
Therefore, we propose an Instance-Aware Cross-Attention module designed to reduce computational overhead while maximally integrating the semantics of instance prompt tokens into the visual tokens. This module is incorporated into the backbone via a residual connection~\cite{he2016deep} and is modulated by a zero-initialized gated parameter ${m_v}$ to ensure training stability.
\noindent\textbf{Attention Masks.}
% intancediffusion怎么说，以及MIGC里面的attention mask
% previous work either rely solely on implicit information in eq（1）to convey the location information，which 对整体的训练提出了挑战，or 将为selfattn计算一个attention mask，但是这
% eq1中的隐式position信息注入对模型的训练提出了巨大的挑战，为了能够高效地，从信息注入效果和运算时间两个层面的高效，将instance-level information注入到
To effectively inject explicit positional information, we employ attention masks $\mathbf{M}\in \mathbb{R}^{F\times N_{ins}\times N_{v}}$ that constrain each instance token to act only on its corresponding visual tokens. In modern deep learning frameworks, this implementation does not significantly increase the forward pass time, even in the absence of FlashAttention~\cite{dao2022flashattention}. The attention mask $\mathbf{\mathrm{M}}$ is defined as follows:
% $$
% M^t_{(i,j)}=\Biggr\{\begin{array}{cc}
%     1 & V^_i \\
%      & 
% \end{array}}
% $$
\[ M^t_{(i, j)} =
\begin{cases}
1 & \text{if } [\, V^t_i, I^t_j \,]=1,\\
-\mathrm{inf} & \text{if } [\, V^t_i, I^t_j \,]=0,
\end{cases} \]
% 其中[\, V^t_i, I^t_j \,]=1表示在frame-t中，和I^t_j对应同一个instance。
Here, $[\, V^t_i, I^t_j \,]=1$ indicates that in frame-$t$, the visual token $V^t_i$ and the instance prompt token $I^t_j$ correspond to the same instance.
% 保证了位置，如何保证visual效果统一呢？需不需要对主体做一个label的融合？

\noindent\textbf{Better Initialization.}
% IMCA的好处不仅在于显式的intance信息注入和高效地计算，并且可以使用对应的DiTBlock中的cross-attention的权重进行初始化，我们发现这可以大幅度地减少模型的训练时间，并且在训练初期通过提供更合理的intance语义信息来使得训练更加稳定。
The advantages of IMCA extend beyond explicit instance information injection and computational efficiency. A key benefit is that its weights can be initialized from the corresponding cross-attention module within the same DiT Block. We find that this initialization strategy substantially reduces the model's training time and enhances stability, particularly in the early stages, by providing more meaningful initial semantic guidance for the instance prompts.

% 增加的MIV-CA模块位于backbone中的self-attention和cross-attention之间，以residual分支的形式，使用一个gated parameter来注入到backbone的visual tokens中。这样在cross-attention模块中，获得语义增强的visual tokens与video caption tokens中的对应部分就可以获得更大的相似度。由于视频整体的语义信息都是由cross-attention module进行注入的，那么通过这种方式就会间接影响cross-attention的信息注入，以达到instance-level的控制。
\noindent\textbf{Overall.}
% backbone中的每个DITBlock都会加入一个IMCA module，which is positioned between the self-attention and cross-attention blocks of the backbone, operating as a residual branch in control of a gated parameter $m_v$ to inject information into the backbone’s visual tokens. In this way, within the cross-attention module, the semantically enhanced visual tokens can achieve higher similarity with the corresponding segments of the video caption tokens,
Each DiT Block within the backbone is augmented with an IMCA module. Positioned between the self-attention and cross-attention blocks, this IMCA module operates as a residual branch modulated by a gated parameter $m_v$, to inject instance-specific information into the backbone’s visual tokens. As a result, when these semantically enhanced visual tokens are processed by the subsequent cross-attention module, they can achieve higher similarity with their corresponding segments in the video caption tokens.
\begin{align}
    \mathbf{V}&=\mathbf{V}+\mathrm{SelfAttn(\mathbf{V})} \\
    \mathbf{V}&=\mathbf{V}+m_v\times \mathrm{IMCA}(\mathbf{V},\mathbf{I},\mathbf{M}) \\
    \mathbf{V}&=\mathbf{V}+\mathrm{CrossAttn(\mathbf{V},\mathbf{T})}
\end{align}
Since the overall semantic information of the video is injected entirely through the cross-attention module, this mechanism indirectly influences the cross-attention’s information injection, thereby enabling instance-level control.

% 需要进行修改
% 这里写的是每个ditblock都加一个ca，但是可以全局加一个ca，然后用类似wan的adalayernorm的方式，这里mt是一个每个ditblock都不同的tensor，然后加上time embedding的偏置项。
\subsection{Shared Timestep-Adaptive Prompt Enhancement Module}
% Shared Timestep-Adaptive Prompt Enhancement (STAPE) Module:

Since our instance prompt tokens are generated by encoding each instance prompt in isolation, they inherently lack global semantics. In contrast, the corresponding content within the global video caption $\mathbf{T} = ({T_1, T_2, \ldots, T_{N_{ctx}}})$, $T_i \in \mathbb{R}^{D_{ctx}}$, contains richer contextual information, including inter-instance interactions and spatial relationships. Consequently, an identical set of instance prompts should yield entirely different outcomes depending on the overarching video caption.

% 由于我们的instance prompt tokens是独立的分隔的instance prompts通过text encode和DiT text embedding layer编码后得到的，所以他们并不具备全局的语义。而global video caption中的对应内容包含了更丰富的全局语义，包括不同intance之间的交互语义、空间位置语义等等。即使是一组完全相同的instance prompts在不同的video captions下的表现可能完全不同。正如前面所讨论的那样，IMCA和CA的关系更像是一种博弈的关系。为了提高训练效率我们freeze了backbone model所以只依赖IMCA来解决这种从局部到全局的语义调整是有些困难的，并且我们通过IMCA注入的instance-level的信息非常的直接和强烈，会导致最终生成的结果可能存在不期望的artifacts。

\noindent\textbf{Instance Prompts Enhancement.}
As previously discussed, the relationship between our IMCA module and the native Cross-Attention resembles an adversarial interplay. This challenge is compounded by our choice to freeze the backbone for training efficiency, which places the entire burden of this local-to-global semantic adjustment solely on the IMCA module. Furthermore, the instance-level information injected by IMCA is highly direct and forceful, which can lead to undesirable artifacts in the final output under certain circumstances.
% 为了解决这个问题，我们提出使用instance prompts enhancement module来为separate instance prompts注入全局的语义。
To address this issue, we propose the Instance Prompts Enhancement module, which is designed to inject global semantics into the separate instance prompts.
% \begin{align}
%     \mathbf{I}&=\mathbf{I}+m_t\times \mathrm{CrossAttn}(\mathbf{I},\mathbf{T})
%     % \mathbf{V}&=\mathbf{V}+m_v\times \mathrm{MIV\_CA}(\mathbf{V},\mathbf{I},\mathbf{M})
% \end{align}

\noindent\textbf{Timestep-Adaptive Modulation.}
% 由于这个instance prompt enhancement module是所有ditblock共享的，类似text encoder和dit的text embedding layer，但是由于不同的ditblock在不同的time step的贡献不是均衡的，因此为了让这个全局的instance prompt enhancement能够timestep-aware的做出adaptive的贡献，我们将residual gated paramter m_t实现为一个D_ins维的tensor实现更加adaptive的block-aware modification，同时复用了ditblock中六个adaptive modulation paramter的第一个param来实现timestep aware的information injection。
The Instance Prompt Enhancement module is shared across all DiT blocks, analogous to the text encoder. However, since the contribution of each DiT block is not uniform across different diffusion timesteps, a static enhancement would be suboptimal. To ensure the module's contribution is both adaptive and timestep-aware, we implement the residual gated parameter $m_t$ as a $D_{ins}$-dimensional tensor, enabling a more fine-grained, adaptive modification. Crucially, to achieve this temporal awareness in a parameter-efficient manner, we compute $\alpha_1$ by repurposing one of the six adaptive modulation parameters from the DiT block's native AdaLN layer~\cite{peebles2023scalable},
\begin{align}
    \mathbf{I}&=\mathbf{I}+(m_t+\alpha_1)\times \mathrm{CrossAttn}(\mathbf{I},\mathbf{T})
    % \mathbf{V}&=\mathbf{V}+m_v\times \mathrm{MIV\_CA}(\mathbf{V},\mathbf{I},\mathbf{M})
\end{align}

\subsection{Spatially-Aware Unconditional Guidance}
% 由于bounding boxes的局限性，某些visual tokens可能对应着不止一个instance token。而经过VAE和DiT backbone的进一步patchify，某些intances可能只对应很少数量的visual token，那么这很可能造成与大instance重合的小instance消失。为了解决这个问题，我们提出了两个方案。
Due to the limitations of bounding boxes, certain visual tokens may correspond to more than one instance token. After further patchification through the VAE and DiT backbone, some instances may correspond to only a small number of visual tokens. This can easily lead to the disappearance of small instances that overlap with larger ones. To address this issue, we further propose a variation of classifier-free-guidance~\cite{ho2022classifier}, termed Spatially-Aware Unconditional Guidance.
\begin{table*}[!t]
\centering
% \resizebox{0.45\textwidth}{!}{%
\resizebox{0.95\textwidth}{!}{%
\begin{tabular}{c|c|cccc|cccccc}
   % \hline
   \toprule
   \multirow{3}{*}{Model}  & \multirow{3}{*}{Resolution} & \multicolumn{4}{c|}{General Metrics} & \multicolumn{6}{c}{Instance-aware metrics} \\
    &&\multirow{2}{*}{FVD$\downarrow$} & \multirow{2}{*}{Temporal$\uparrow$} & \multirow{2}{*}{CLIP-B$\uparrow$} & \multirow{2}{*}{CLIP-L$\uparrow$} & \multicolumn{3}{c}{Area-Weighted Mean} & \multicolumn{3}{c}{Mean}  \\
   &&&&&& IoU\%$\uparrow$&CLIP-B$\uparrow$&CLIP-L$\uparrow$ &IoU\%$\uparrow$&CLIP-B$\uparrow$&CLIP-L$\uparrow$ \\
   % \hline
   \midrule
   Wan2.1&480P& 2837.22 & 0.9834&0.3088 & 0.2480& 0.2721&0.2172& 0.1625&0.1454&0.2179&0.1635 \\
   Wan2.2&720P& 2687.98 &0.9867 &0.3110 &0.2495 &0.2717 &0.2207&0.1698 &0.1547&0.2198&0.1720\\
   HunyuanVideo&720P&2781.50&0.9808&0.2977&0.2395& 0.3172 &0.2178&0.1674& 0.1543&0.2230& 0.1700\\
   CogVideoX&480P&3416.78&\textbf{0.9939}&0.3002&0.2483&0.2564 &0.2138&0.1594 &0.1339&0.2187&0.1635\\
   SkyReaelsV2&540P&3155.76&0.9797&0.2965& 0.2461&0.2478 &0.2154& 0.1611 &0.1386&0.2176&0.1631\\
   % \hline
   \midrule
   \multirow{2}{*}{\textbf{InstanceV(ours)}}&480P&2313.06&0.9906&0.3123&0.2519& \textbf{0.7298}&0.2321&0.1793 &\textbf{0.6090}&\textbf{0.2322}&0.1782\\
   &720P&\textbf{2285.22}&0.9924&\textbf{0.3167}&\textbf{0.2530}& 0.7228&\textbf{0.2324}& \textbf{0.1838} &0.5910&0.2313&\textbf{0.1849}\\
   % \hline
   \bottomrule
\end{tabular}%
}
\caption{Quantitative comparisons. 
% 我们主要report了proposed instwan framework with opensource sota t2v video diffusion models，including 480P and 720P。from two perspectives including general video quality metrics and our proposed instancebench。
We primarily report the results of the proposed InstanceV framework integrated with open-source state-of-the-art text-to-video diffusion models, evaluated at both 480P and 720P resolutions.
The evaluation is conducted from two perspectives: (1) general video quality metrics, and (2) our proposed InstanceBench for instance-level assessment.
}
\label{tab:results0}
% \end{table}
\vspace{-10pt}
\end{table*}
% }
% \vspace{-1em} 
% \subsubsection{Instance Prompt Classifier-Free Guidance}
% 虽然float-version的attention mask可以很好地解决小instance消失的问题，但是float版本的attention mask会实际占用更大的显存，并一定程度上增加整体计算量。因此我们还提出了一种training free的方法，即Instance Prompts Classifier-free guidance。
% 在unconditional的前向过程中，依然使用attention masks进行显式的位置信息注入，但是留空对应的instance prompts。这会给cross attention模块注入位置信息，留语义可以让cross attention模块更好地从全局出发把握不同intances之间的前后层级信息，避免large instances的过强的平面化的语义干涉，从而更好的生成smaller instances。
% Although the floating-point attention mask effectively addresses the disappearance of small instances, it incurs higher memory consumption and increases overall computational cost. To mitigate this, we further propose a training-free approach, termed Instance-Prompts Classifier-Free Guidance~\cite{ho2022classifier}.

During the unconditional forward pass, the attention masks are still applied to inject explicit spatial information, while the corresponding instance prompts are left empty,
\begin{align}
    \tilde{\boldsymbol{\epsilon}_\theta}(\mathbf{z},\mathbf{I},\mathbf{M})=(1+w)\boldsymbol{\epsilon}_\theta(\mathbf{z},\mathbf{I},\mathbf{M})-w\boldsymbol{\epsilon}_\theta(\mathbf{z},\mathbf{M}).
\end{align}
This setup allows the cross-attention module to receive positional cues while leaving semantic content unset. As a result, the cross-attention module can better reason about hierarchical relationships among different instances from a global perspective, reducing the overly flattened semantic interference from large instances and thereby improving the generation of smaller instances.
% 具体实现过程中，对于留空的instance prompts，我们复用了text encoder中<extra_id>，相较于新增加额外的tokens，经过预训练的<extra_id>显然拥有更好的非明确语义。
In implementation, for the empty instance prompts, we reuse the \textless extra\_id\textgreater{} tokens from the text encoder. Compared with introducing new tokens, the pretrained \textless extra\_id\textgreater{} tokens naturally possess richer and more general semantic representations.
% 图

\begin{figure*}[t]
    \centering
    \includegraphics[width=\linewidth]{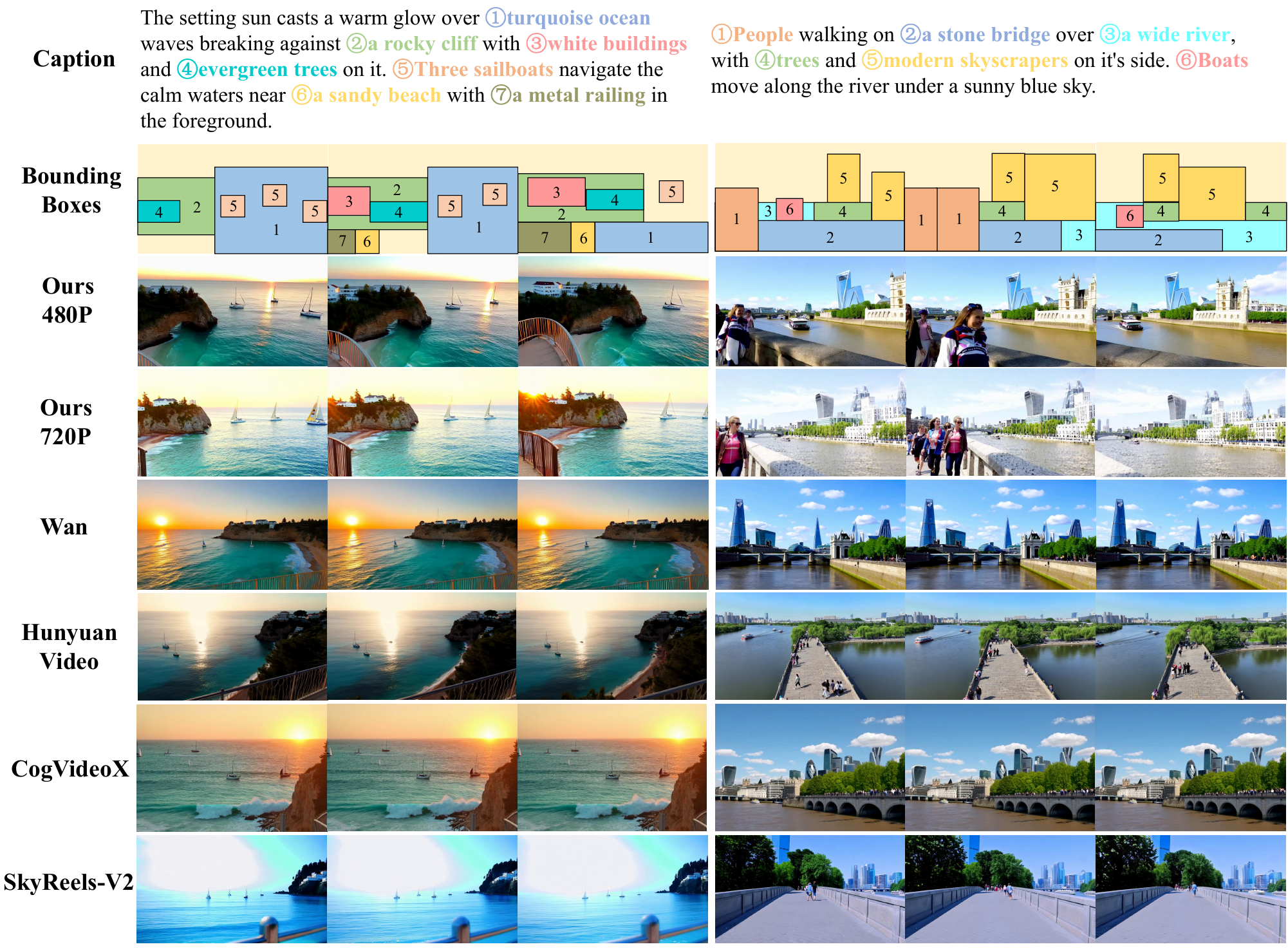}
    \caption{
Quality comparison between our proposed InstanceV and state-of-the-art video diffusion models. The first, middle, and last frames are shown for illustration.
    }
    \label{fig:quality}
    \vspace{-5pt}
\end{figure*}

\section{Experiments}
\subsection{Implementation Details}
\subsubsection{Baselines}
% 我们选择流行的Wan-Series模型作为我们的backbone，训练过程中backbone是frozen的。除了Wan，我们还和HunyuanVideo以及CogvideoX等开源sota模型进行了对比，同时我们也和闭源商业模型进行了对比。对于每个模型我们生成了300个进行评测。

We compare our model against state-of-the-art methods, including Wan, HunyuanVideo, CogVideoX, and SkyReelsV2.
Since InstanceCap has not released any fine-tuned models, we implement a substitute version as described in the supplementary material.
Following Figure.~\ref{fig:fig0}, we generate a total of 100 videos for each model in 5 groups, including   
(1) main subjects,
(2) dense small objects,
(3) complex scene settings,
(4) trajectory control, and
(5) camera control.

\subsection{Evaluation Metrics}
% 除了通用的评价指标FVD以及通用benchmark，如vbench，vbench2等。针对instance-level video generation任务的特殊性，我们提出了一个包含两个protocol的instancebench。对于生成的视频我们使用data preparation方法得到标注。（1）对于label相同的instance，我们计算bounding box的IoU。（2）对于生成的视频，我们选择bounding box中的图片，计算其与对应instance label的clip-score。InstanceBench的两个protocol分别针对layout和visual进行了评测。
% 在实际实现中我们发现，对不同大小的instance的IoU和ClipScore进行取平均计算作为整个video的IoU和ClipScore在某种程度上是不公平的。因为面积更大的instance显然包含更多的visual语义和对应的video caption中的text语义，对视频的视觉效果的影响更大。因此，除了report上述一般的IoU和ClipScore之外，我们还评估了area-weighted的IoU和ClipScore。

% In addition to general evaluation metrics such as FVD, temporal consistency, text-to-video alignment and widely used benchmarks like VBench~\cite{huang2024vbench} a
nd VBench2, 
We design a comprehensive InstanceBench for instance-aware evaluation of video diffusion models.
InstanceBench consists of two sets of metrics. For general-purpose evaluation, we include FVD~\cite{unterthiner2018towards}, Temporal~\cite{huang2024vbench}, and CLIPScore metrics.
% The annotations for generated videos are obtained using our data preparation pipeline.
% 假设对于real video和generated video，使用~\ref{sec:data_preparation}中的pipeline分别得到了一组instance information。
For instance-aware evaluation, we design two complementary testing protocols,
supposing that a set of instance information is obtained using the pipeline described in Section~\ref{sec:data_preparation} for both real and generated videos, ,
\textbf{(1)} For instances sharing the same label, we compute the bounding box IoU~\cite{rezatofighi2019generalized} to measure spatial alignment and layout consistency.
\textbf{(2)} We extract from generated videos the image patches within the ground-truth bounding boxes and compute their CLIP-Score with the corresponding instance prompts to evaluate visual-semantic fidelity.
Together, these two protocols provide a joint evaluation of layout accuracy and instance-level visual quality.
However, simply averaging the IoU and CLIP-Score across instances of different sizes can be unfair due to their varying degrees of influence on the generated video.
% Larger instances naturally carry more visual semantics and correspond to richer textual descriptions in the video caption, thus exerting a stronger influence on the overall perceptual quality of the video.
To address this, we additionally evaluate area-weighted IoU and area-weighted CLIP-Score, which better reflect the semantic and visual contributions of different instances.

\subsection{Comparison Results}
\subsubsection{Qualitative comparisons}

As show in Figure, InstanceV achieves accurate instance-level control conditioned on the additional grounding information. While maintaining the overall visual quality of the backbone, it further enhances the fidelity of instance-specific details. Moreover, both inter-instance relations and the consistency between local instances and global semantics are well preserved. In addition, control conditions that are difficult to express purely through text, such as camera motion, can also be indirectly achieved via instance-level grounding information. Without additional instance-level grounding information, different models tend to produce visually diverse videos even when conditioned on the same caption. We observe that existing state-of-the-art models often fail to accurately represent instance-level semantics, trapping themselves in missing or misattributed instances even explicitly described in the caption. Moreover, although the captions may clearly describe motion, the corresponding instances are sometimes generated too small or too blurred, resulting in videos that appear overly static.
Specifically, outputs from the Wan-series and HunyuanVideo models may contain noticeable artifacts under more complex semantics. While HunyuanVideo produces videos with richer motion dynamics, this often comes at the expense of temporal consistency. In contrast, CogVideoX frequently generates nearly static videos, lacking sufficient motion cues and resulting in visually stagnant scenes. For SkyReelV2, videos with moderately complex motion often exhibit frame-to-frame flickering and suffer from poor temporal coherence.

\subsubsection{Quantitative Comparisons}
% 从具体的数值指标可以看出，FVD方面我们的instwan框架提升非常明显，表示在对prompt所要表达的视觉语义的还原上，有着显著的提升。同时，text-to-video alignment也有一定的提升，表示在增加instance-level的grounding 信息后，有助于生成更符合文本语义的视频。
From the quantitative results, our InstanceV framework shows a notable improvement in FVD, indicating a significant enhancement in faithfully restoring the visual semantics expressed by the prompts.
Meanwhile, the text-to-video alignment also improves, suggesting that incorporating instance-level grounding information helps the model generate videos that more accurately correspond to the intended textual semantics.
% 尽管在temporal consistency上，instwan的指标略低于cogvideox，但是我们发现这是因为cogvideox倾向于生成非常静止，缺少动态元素的视频。而我们的instwan在instance-level grouding information的帮助下，可以为不同instances乃至全局语义都赋予合理的动态范围。
Although InstanceV shows slightly lower scores in temporal consistency compared to CogVideoX, we observe that this is mainly because CogVideoX tends to generate overly static videos with limited motion. 
% In contrast, benefiting from the instance-level grounding information, InstWan introduces a more reasonable and diverse range of dynamics across both individual instances and global semantics, leading to videos with more natural temporal variations.
% 在InstanceBench上，我们提出的InstWan框架同样提升明显。从IoU指标可以明显看出，增加intance-prompt到bounding-box的grounding information可以显著改善生成视频的layout信息。在面积加权平局的IoU指标下，这个提升更加明显。
Our proposed InstanceV framework also demonstrates significant improvements when it comes to instance-aware metrics.
From the IoU metric, it is evident that incorporating additional instance-level grounding information substantially enhances the layout quality of the generated videos.
This improvement becomes even more pronounced when evaluated using the area-weighted IoU, which better reflects the influence of larger instances on the overall visual structure. 
% 即使480P-version的instwan在IoU指标上略优于720P-version，但是在CLIPScore指标上720P-version的instwan优于480Pversion，得益于更高的分辨率和更加清晰的局部细节。
Although the 480P version of InstanceV achieves slightly higher IoU scores than the 720P version, the 720P version outperforms in terms of CLIPScore, benefiting from its higher resolution.

% \begin{figure*}[!t]
%     \centering
%     \includegraphics[width=0.99\textwidth]{figures/ablation1.pdf}
%     \caption{ablation}
%     \label{fig:ablation1}
% \end{figure*}

% \begin{figure}[!t]
%     \centering
%     \includegraphics[width=\linewidth]{figures/ablation2.pdf}
%     \caption{Ablation on Instance-aware Masked Cross-Attention}
%     \label{fig:miv}
% \end{figure}

\begin{figure*}[!t]
    \centering
    \includegraphics[width=\linewidth]{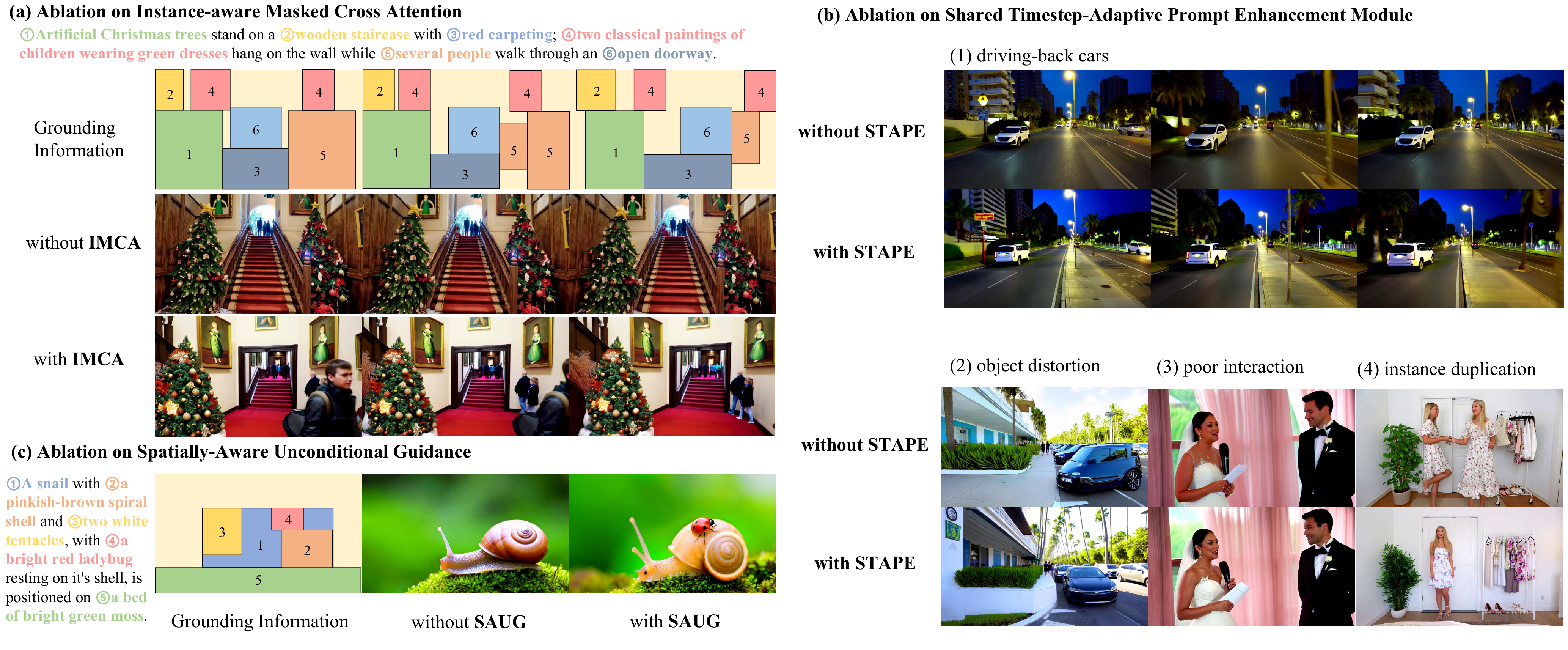}
    \caption{Ablation studies on the proposed modules.}
    \label{fig:ablation}
    \vspace{-5pt}
\end{figure*}

\begin{table}[t]
    \centering
    \resizebox{1.0\linewidth}{!}{
    \begin{tabular}{cccc|cccc}
    \toprule
        \multirow{2}{*}{Backbone} & \multirow{2}{*}{IMCA} &\multirow{2}{*}{STAPE}&\multirow{2}{*}{SAUG}&\multirow{2}{*}{Temporal}&\multicolumn{3}{c}{Area-Weighted Mean}\\
         &&&&&IoU\%$\uparrow$&CLIP-B\%$\uparrow$&CLIP-L\%$\uparrow$ \\
        \midrule
        $\checkmark$ &&&&0.9834&0.2721&0.2172&0.1625 \\
        $\checkmark$ &$\checkmark$&&&0.9885&0.7157&0.2315&0.1781 \\
        $\checkmark$ &$\checkmark$&$\checkmark$&&0.9903&0.7082&0.2303&0.1766 \\
        $\checkmark$ &$\checkmark$&$\checkmark$&$\checkmark$&\textbf{0.9906}&\textbf{0.7298}&\textbf{0.2321}&\textbf{0.1793} \\
    \bottomrule
    \end{tabular}}
    \caption{Quantitative Ablation Studies.}
    \label{tab:ablation}
    \vspace{-5pt}
\end{table}

\subsection{Ablation Studies}
We conduct ablation studies on three variants:
(1) Instance-aware Masked Cross-Attention (2) Shared Timestep-Adaptive Prompt Enhancement Module (3) Spatially-Aware Unconditional Guidance.
% 从fig4中可以看到，masked instance visual cross attention在backbone的基础上决定了视频整体的layout和动态范围。在复杂的语义下可以帮助模型更好的理解video caption中不同instance的特征属性和空间位置，赋予不同instances之间更加合理的交互和关联，缓解artifacts的产生、避免完全静止视频的生成。尽管instance的视觉语义与instance prompt比较接近，但是由于instance prompts的作用域是比较有限的，加之后续cross attention模块对全局的影响，最终生成的结果中很有可能产生artifacts，（1）倒行的车辆和人导致的语义错误（2）instance形象的畸形导致的视觉语义错误 （3）不同instances之前失去原本的交互语义而造成的类似“断连”的artifact（4）由于cross attention所理解的instance位置与instance bounding boxes差异较大的，有可能产生两个重复的物体所导致的text-visual语义不一致性和空间不一致性。（5）同时由于可能的大instance的存在所导致的小instance的消失的artifact的grounding information-visual的语义不一致。当加入enhanced instances prompts时，这些问题可以得到有效解决。这是因为instance prompts通过与全局的video caption交互，将独立的instance prompts与全局语义联合起来，某种程度上与cross attention module达成共识，通过影响instances的属性来达到更加合理的全局语义。fig4和supplementary中的示例可以非常形象的展示这一点。对于（5），我们所提出的两种独立的策略，float-version attention mask和instance prompts classifier-free guidance都可以很好地解决这一点。在实际训练和推理过程中，前者需要的显存显著增加，而后者则需要更长的训练更新步数。从效果上来讲，前者的控制效果更好，但是有可能稍微影响视频的时序一致性；而后者的影响则温和一些，但是在某些非常复杂的语义条件下可能失效。我们在补充材料中进行了更为详细的讨论。
As shown in Fig.~\ref{fig:ablation}(a), the Instance-aware Maksed Cross-Attention module plays a crucial role in determining the overall video layout and motion range. Under complex semantics, it helps the model better understand the feature attributes and spatial positions of different instances described in the overall video caption. This mechanism further enables more reasonable interactions and associations among instances, thereby alleviating artifacts and preventing the generation of completely static videos.
Due to the limited influence of mutually independent instance prompts on the overall semantics, various types of artifacts occur, as shown in Fig.~\ref{fig:ablation}(b), including:
(1) semantic errors caused by reversed motions of vehicles or humans;
(2) visual semantic distortions due to malformed instance appearances;
(3) disconnection artifacts, where interactions between different instances are lost;
(4) duplicated instances caused by disagreements between IMCA and native cross-attention.
With the help of SPAPE, these issues are effectively alleviated, leading to improved overall temporal consistency and text-visual alignment, as shown in Table~\ref{tab:ablation}.
Benefiting from proposed SAUG, the generation of small objects is significantly improved. We observe increases in IoU and CLIPScore as shown in Table.~\ref{tab:ablation}, indicating that these small objects are no longer overlooked and can meaningfully contribute to the overall semantic of the generated video.
\section{Conclusion}
In this work, we propose InstanceV, a novel framework for instance-level controllable video generation.
By leveraging Instance-aware Masked Cross-Attention mechanism, InstanceV is able to generate instances with precise spatial localization and well-defined attributes.
The proposed Shared Timestep-Adaptive Prompt Enhancement Module improves the coherence between local instance semantics and global video context in a parameter-efficient manner.
In addition, a Spatially-Aware Unconditional Guidance strategy is incorporated during both training and inference to further refine the details and fidelity of small instances.
Extensive experiments demonstrate that InstanceV achieves not only accurate instance-level control, but also stronger alignment with video captions and richer instance details, jointly contributing to a significant improvement in the overall quality of generated videos.
% \section{Final copy}

% You must include your signed IEEE copyright release form when you submit your finished paper.
% We MUST have this form before your paper can be published in the proceedings.

% Please direct any questions to the production editor in charge of these proceedings at the IEEE Computer Society Press:
% \url{https://www.computer.org/about/contact}.
% {
    \small
    \bibliographystyle{ieeenat_fullname}
    \bibliography{main}
% }
% \input{sec/2_formatting}
% \clearpage
% \setcounter{page}{1}
% \maketitlesupplementary
\section{Overview}
The supplementary material is composed of:
\begin{itemize}
    \item Computational efficiency analysis (Sec.~\ref{sec:computation})
    % \item More related works and discussion on video diffusion (Sec.~\ref{sec:related})
    \item Implementation details and more comparison results. (Sec.~\ref{sec:impl})
    \item More details and discussion on model implementation (Sec.~\ref{sec:model})
    \item Potential limitations and future work (Sec.~\ref{sec:model})
\end{itemize}

\section{Computational Efficiency Analysis}
\label{sec:computation}
% % 
% Having the supplementary compiled together with the main paper means that:
% % 
% \begin{itemize}
% \item The supplementary can back-reference sections of the main paper, for example, we can refer to \cref{sec:intro};
% \item The main paper can forward reference sub-sections within the supplementary explicitly (e.g. referring to a particular experiment); 
% \item When submitted to arXiv, the supplementary will already included at the end of the paper.
% \end{itemize}
% % 
% To split the supplementary pages from the main paper, you can use \href{https://support.apple.com/en-ca/guide/preview/prvw11793/mac#:~:text=Delete%20a%20page%20from%20a,or%20choose%20Edit%20%3E%20Delete).}{Preview (on macOS)}, \href{https://www.adobe.com/acrobat/how-to/delete-pages-from-pdf.html#:~:text=Choose%20%E2%80%9CTools%E2%80%9D%20%3E%20%E2%80%9COrganize,or%20pages%20from%20the%20file.}{Adobe Acrobat} (on all OSs), as well as \href{https://superuser.com/questions/517986/is-it-possible-to-delete-some-pages-of-a-pdf-document}{command line tools}.

% intro第二段的方法对比图（学习一下ppt画图）
% controllable video diffusion -- 可以再补充一些论文 （开始关注组）
% sec3 补充一些例子
% lightweight 表格说明
% expr implementation detail和平均metric
% more about model arch （1）frame dim 折叠 （2）cfg extra_id（3）eil的实现
% human study
% instancecap

% optional 
% cross attn 分析，layout形成的时间节点
% 对attention map的实现，统计量
% 可以加一个float attention mask的说明
% 更多的video metric，可以弄成一个仓库方便后续使用

% AI instance generation and results

% 正如table1所示，我们提出的instance-level控制框架的参数量和计算效率都要显著高于之前的方法，值得一提的时，instancegen并没有公开代码，因此table1中的结果是我们理论分析的结果。
% 我们列举出了生成一个480P的视频，每个DiTBlock的不同模块的FLOPS，注意，我们在table1中report的added flops是对整个vdm backbone而言的，所以根据下表计算出的added flops的结果可能和table1中有所出入。

% 更多对比细节，我们的added params的backbone是相对这个vdm backbone而言的，在不同实现下，不只包括ditblocks，还包括可能存在的time-embeddings layers和text embeddings layer以及video token的patchifier。
% added time costs我们是取生成一百个包含5个instance的image/video的平均added time而言的。

As shown in Table.1 in main text, our proposed instance-level control framework \textbf{InstanceV} achieves \textit{significantly lower parameter overhead} and \textit{higher computational efficiency} compared to previous approaches. It is worth noting that InstanceGen~\cite{sella2025instancegen} \textit{has not released} its official implementation, and thus the results reported for InstanceGen in Table.1 of the main text are based on our theoretical analysis.

More Comparison Details:
\textbf{(1)} The reported added parameters are measured relative to the video diffusion model backbone, which may include not only DiT blocks but also additional components such as timestep embedding layers, text embedding layers, and patchifying modules for video tokenization, depending on the specific implementation.
\textbf{(2)} The added time cost is computed as the \textit{average extra} inference time required to generate \textbf{100} samples, grounded additionally by \textbf{5}  instance-level annotations. We constrain the number of instances because, for some existing approaches~\cite{wang2024instancediffusion}, the instance count can directly affect the required number of \textit{denoising steps}, which would otherwise lead to unfair comparisons.

We provide the FLOPs of each component within a single DiT block for generating a 480P video. 
Note that the added FLOPs reported in Table.1 of the main text correspond to the entire VDM backbone. 
Therefore, the FLOPs estimated from the table below may slightly differ from the results reported in Table.1 of the main text due to implementation variations and additional architectural components.

% \begin{table}[]
%     \centering
%     \begin{tabular}{c|c}
%          &  \\
%          & 
%     \end{tabular}
%     \caption{Caption}
%     \label{tab:placeholder}
% \end{table}

% \section{More Related Works on Video Diffusion}
% \label{sec:related}
\section{Implementation Details}
\label{sec:impl}
\subsection{Metrics}
% for general purpose,我们主要选择了三组指标来衡量视频质量：
\noindent\textbf{General Metrics.}
\begin{itemize}
    \item \textbf{FVD}
    Fréchet Video Distance (FVD)~\cite{unterthiner2019fvd} is a widely adopted metric for evaluating the quality of generated videos. Similar to the Fréchet Inception Distance (FID) used in image generation, FVD measures the distributional distance between real and generated video feature representations extracted by a pretrained spatiotemporal network (typically an I3D model~\cite{carreira2017quo}). It jointly accounts for both visual fidelity and temporal consistency by capturing correlations across consecutive frames. Lower FVD values indicate that the generated videos are closer to real ones in terms of appearance and motion dynamics, and thus represent better overall generation quality.
    \item \textbf{Temporal}
    In the VBench~\cite{huang2024vbench} benchmark suite, one of the key dimensions under Temporal Quality is \textit{Motion Smoothness}. Rather than measuring an abstract “temporal score,” VBench quantifies how physically plausible and fluid the motion in generated videos is: it compares the actual frames with interpolated frames (via a frame-interpolation model) and computes discrepancies to reflect coherence and naturalness of motion. A higher motion smoothness score indicates that the generated video exhibits more stable and realistic motion over time, with fewer sudden jitters or unnatural transitions. 
    \item \textbf{CLIP-Score}
    CLIP Score~\cite{hessel2021clipscore} leverages the pretrained CLIP model~\cite{radfordLearningTransferableVisual2021a} to evaluate the semantic alignment between generated visual content and the corresponding text descriptions. Specifically, both the image and its associated prompt are encoded into a shared multimodal embedding space, and their similarity is measured using cosine similarity. A higher CLIP Score indicates stronger semantic consistency between the visual appearance and the textual intent. Since CLIP is trained on a large corpus of diverse image–text pairs, this metric provides a robust and scalable evaluation of high-level semantic relevance in generative tasks. 
    % 对于视频我们计算每一帧的clip score然后取平均作为视频-level的clip score
    \textit{For videos, we compute the CLIP score for each individual frame and then take the average across all frames to obtain a video-level CLIP score.}
\end{itemize}

\noindent\textbf{Instance-aware Metrics}
\begin{itemize}
    \item IoU
Intersection over Union (IoU)~\cite{rezatofighi2019generalized}, also known as the Jaccard Index, is a widely used metric for evaluating the overlap between two sets, commonly applied in object detection and segmentation tasks. 
Given a predicted region $A$ and a ground-truth region $B$, the IoU is defined as the ratio of the area of their intersection to the area of their union:

\[
\text{IoU}(A, B) = \frac{|A \cap B|}{|A \cup B|} = \frac{|A \cap B|}{|A| + |B| - |A \cap B|}.
\]

IoU values range from 0 to 1, where 0 indicates no overlap and 1 indicates perfect alignment. 
In video or instance-level tasks, IoU can be computed per frame or per instance, and averaged to assess temporal or multi-object consistency. 
% Table.~\ref{tab:results0}和Table.~\ref{tab:ablation}中的表格的CLIP-B和CLIP-L指标分别是使用CLIP-base和CLIP-Large模型进行计算的。
% instance-aware metrics 中的IoU的定义和上述一致，如果某个instance在生成视频中没有出现时，其IoU计为0.
The CLIP-B and CLIP-L metrics reported in Table.2 and Table.3 of the main text are computed using the CLIP-ViT-B/32 and CLIP-ViT-L/14 models, respectively.
The definition of IoU in the instance-aware metrics is consistent with the above; \textit{if a specific instance does not appear in the generated video, its IoU is assigned a value of 0.}
\item Instance-level CLIP-Score
% instance-level的clip-Score在计算方式上和一般的CLIP-score没有任何区别，这里我们根据bounding-boxes对图片进行裁剪，然后resize到指定大小后然计算CLIP-Score。如果某个instance在生成视频中没有出现，那么其CLIP-Score记为0.
The instance-level CLIP-Score is computed in the same manner as the standard CLIP-Score. Specifically, each instance is first cropped from the image according to its bounding box and then resized to the required input size before computing the CLIP-Score. \textit{If a particular instance does not appear in the generated video, its CLIP-Score is set to 0}.
\end{itemize}

\begin{table*}[!t]
    \centering
    \resizebox{\linewidth}{!}{
    \begin{tabular}{c|c|ccccc}
    \toprule
    \textbf{Methods} & \textbf{Resolution}    & \textbf{Visual Quality} $\uparrow$ &\textbf{Temporal Consistency}$\uparrow$&\textbf{Dynamic Degree}$\uparrow$&\textbf{Text-to-Video Alignment}$\uparrow$&\textbf{Factual Consistency}$\uparrow$ \\
     \midrule    
    Wan2.1~\cite{wan2025wan}  &  480P  & 2.99355 & 3.14705 & 3.22359 & 3.40019 & 2.96861\\
    Wan2.2   & 720P &2.93241 & 3.11781 & 3.00619 &  3.25030  &   2.96679\\
    HunyuanVideo~\cite{kong2024hunyuanvideo}   & 720P &2.84249 & 3.02305 & 2.89554 &    3.14622  &  2.88761\\
    CogVideoX~\cite{yang2024cogvideox}    &480P &2.61276 &  2.79843 & 2.84793 & 3.05415 & 2.62031\\
     SkyReelsV2~\cite{chen2025skyreels}  & 540P &2.62439 & 2.81272 &    2.76471 & 3.05063 &  2.63757\\
\midrule
    \multirow{2}{*}{\textbf{InstanceV(ours)}} &480P&\textbf{3.10872} &   3.16971 & \textbf{3.42722} &    \textbf{3.47803} & 3.05342  \\
    &720P& 3.03335 & \textbf{3.23439} & 3.20825 & 3.33263 & \textbf{3.11876}  \\
      \bottomrule   
    \end{tabular}
    }
    \caption{
Quantitative comparison of our proposed InstanceV with state-of-the-art video diffusion models using \textit{VideoScore}.
    }
    \label{tab:vs}
% \end{table*}
\vspace{0.5cm}
% \begin{table*}[t]
    \centering
    \resizebox{\linewidth}{!}{
    \begin{tabular}{c|c|ccccc}
    \toprule
    \textbf{Methods} & \textbf{Resolution}    & \textbf{Visual Quality} $\uparrow$ &\textbf{Temporal Consistency}$\uparrow$&\textbf{Dynamic Degree}$\uparrow$&\textbf{Text-to-Video Alignment}$\uparrow$&\textbf{Factual Consistency}$\uparrow$ \\
     \midrule    
    Wan2.1  &  480P  & 2.98814 &   2.87894 &   3.15407 & 3.19866 &   2.79638\\
    Wan2.2   & 720P &3.22745 & 3.20388 &   3.12385 & 3.14581 & 3.18957\\
    HunyuanVideo   & 720P &2.99697 & 2.99133 &2.88472  &  2.91647& 2.99167\\
    CogVideoX    &480P &2.67665 & 2.69979& 2.71733 & 2.78533& 2.61382\\
     SkyReelsV2  & 540P &2.63698 &   2.68429& 2.59490   &  2.75576  &  2.59753\\
\midrule
        \multirow{2}{*}{\textbf{InstanceV(ours)}} &480P& 3.33516 & 3.24925 & \textbf{3.52941} & \textbf{3.43886} & 3.22753 \\
    &720P&\textbf{3.49686} & \textbf{3.44247} & 3.49406 & 3.35064 & \textbf{3.45481}\\
      \bottomrule   
    \end{tabular}
    }
    \caption{Quantitative comparison of our proposed InstanceV with state-of-the-art video diffusion models using \textit{VideoScore-V1.1}.}
    \label{tab:vs1.1}
\end{table*}

\subsection{Experiment settings.}
% 在训练过程中，我们选择了batchsize=2，optimizer选择，精度选择了bfloat16，实验过程中我们发现由于bfloat16的精度限制、lr大小和参数本身的大小和grad导致某些added-param的层的参数在更新到一些特定的值之后停止更新。因此我们选择了使用一个阶段性的lr rate策略，（1）前1000step我们采用1e-5的lr rate来确保训练初期的稳定性，（2）1000到3000step我们让lr rate线性增长到5e-4，（3）维持5e-4的学习率训练到8000step，（4）然后让lr rate线性降低到1e-5训练到10000step，保证训练后期参数收敛的稳定性。
\noindent\textbf{Training Details.}
During training, we adopt a batch size of 2 and use the Adam optimizer with bfloat16 precision. 
% For the proposed SAUG strategies, we choose 20\% for the unconditional forward probability during training.
For the proposed SAUG strategy, we adopt an unconditional forward pass probability of \textbf{20\%} during training, following the standard practice in classifier-free guidance~\cite{ho2022classifier}.
We observed that due to the limited numerical precision of bfloat16, the learning rate, as well as the magnitude of certain added-parameter layers and their gradients, some parameters occasionally ceased updating after reaching specific values. 
To address this issue, we employ a staged learning rate schedule: 
\textbf{(1)} for the first 1,000 steps, a learning rate of 1e-5 is used to ensure stability at the early stage of training; 
\textbf{(2)} from step 1,000 to 3,000, the learning rate is linearly increased to 5e-4; 
\textbf{(3)} the learning rate is maintained at 5e-4 until step 8,000; 
\textbf{(4)} finally, it is linearly decreased back to 1e-5 until step 10,000 to guarantee stable convergence in the later stages of training.

% 可以选两个对比图，对于cfg的选择，
\noindent\textbf{Testing Details.}

\begin{figure*}[t]
    \centering
    \includegraphics[width=\linewidth]{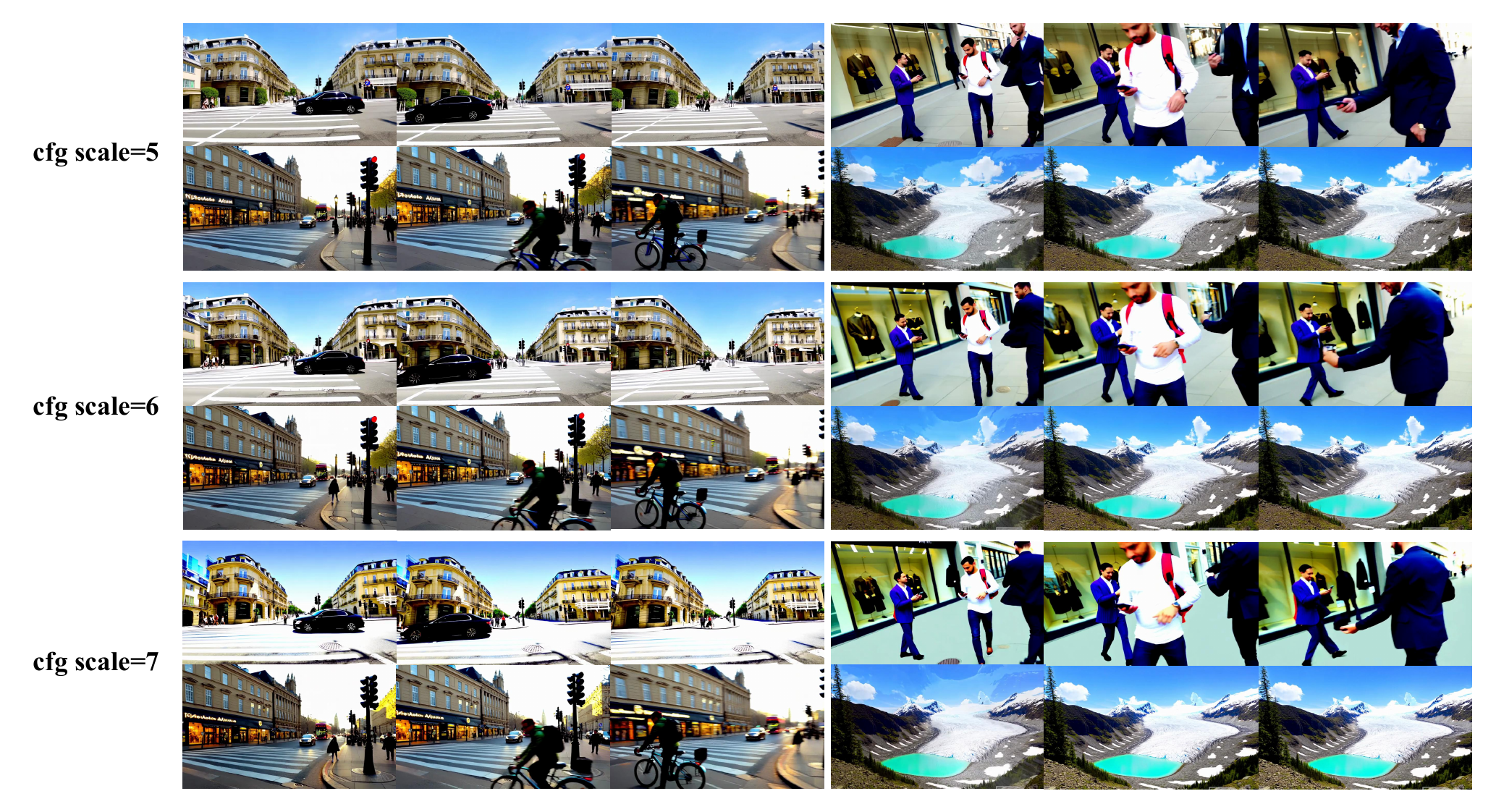}
    \caption{
    % ablation on cfg scale。对于cfg scale=5,6,7，我们分别选择了四个视频进行展示，每个视频选择了第一帧，中间帧和末尾帧。
    Ablation on CFG scale. We compare three settings (CFG = 5, 6, 7) by visualizing four representative videos, each shown with its first frame, a middle frame, and the final frame.
    }
    \label{fig:cfg}
\end{figure*}

\noindent\textbf{CFG.}
% 在Inference时，我们选择cfg scale=5，which is same as general T2V task of backbone.
During inference, \textit{we set the CFG scale to \textbf{5}}, which is identical to the setting used for the backbone’s general text-to-video generation task.
% 对于added module，unconditional forward时，所有added module的输入全为全0tensor。
For the added modules, \textit{all additional instance-level grounding inputs are replaced with all-zero tensors during unconditional forward passes}.
% 不同cfg scale对最终生成视频结果的影响效果可以见图
The influence of different CFG scales on the final generated videos is illustrated in Figure.~\ref{fig:cfg}.
Since the backbone uses a CFG scale of 5 for its original text-to-video task during inference, \textit{setting the CFG scale below 5 leads to two issues}: \textbf{(1)} the additional instance-level grounding information has only a minimal effect, and \textbf{(2)} the text–visual alignment becomes noticeably weaker. When the CFG scale is increased beyond 5, we observe that although the model can, in some cases, better follow the instance-level grounding inputs, the overall visual quality of the generated videos slightly degrades. For example, the contrast becomes overly high and certain colors are lost, which noticeably reduces the distinguishability between similar hues and causes the boundaries between different instances to appear blurred. Considering this trade-off, we therefore adopt a CFG scale of 5 as our default setting.

% 由于backbone inference时原始T2V任务的cfg就是5，所以当cfg-scale小于5时，不仅additional instance-level grounding information起到的作用非常有限，text-visual alignment也会变差。当cfg scale大于5时，我们发现虽然某些情况下对instance-level grounding information可以follow得更好，但是我们发现视频的图像质量上有略微的下降，具体表现在对比度变高和色彩的丢失。因此权衡不同方面的效果，我们最终选择了cfg-scale=5.

\subsection{More comparisons}

\begin{figure*}[t]
    \centering
    \includegraphics[width=\linewidth]{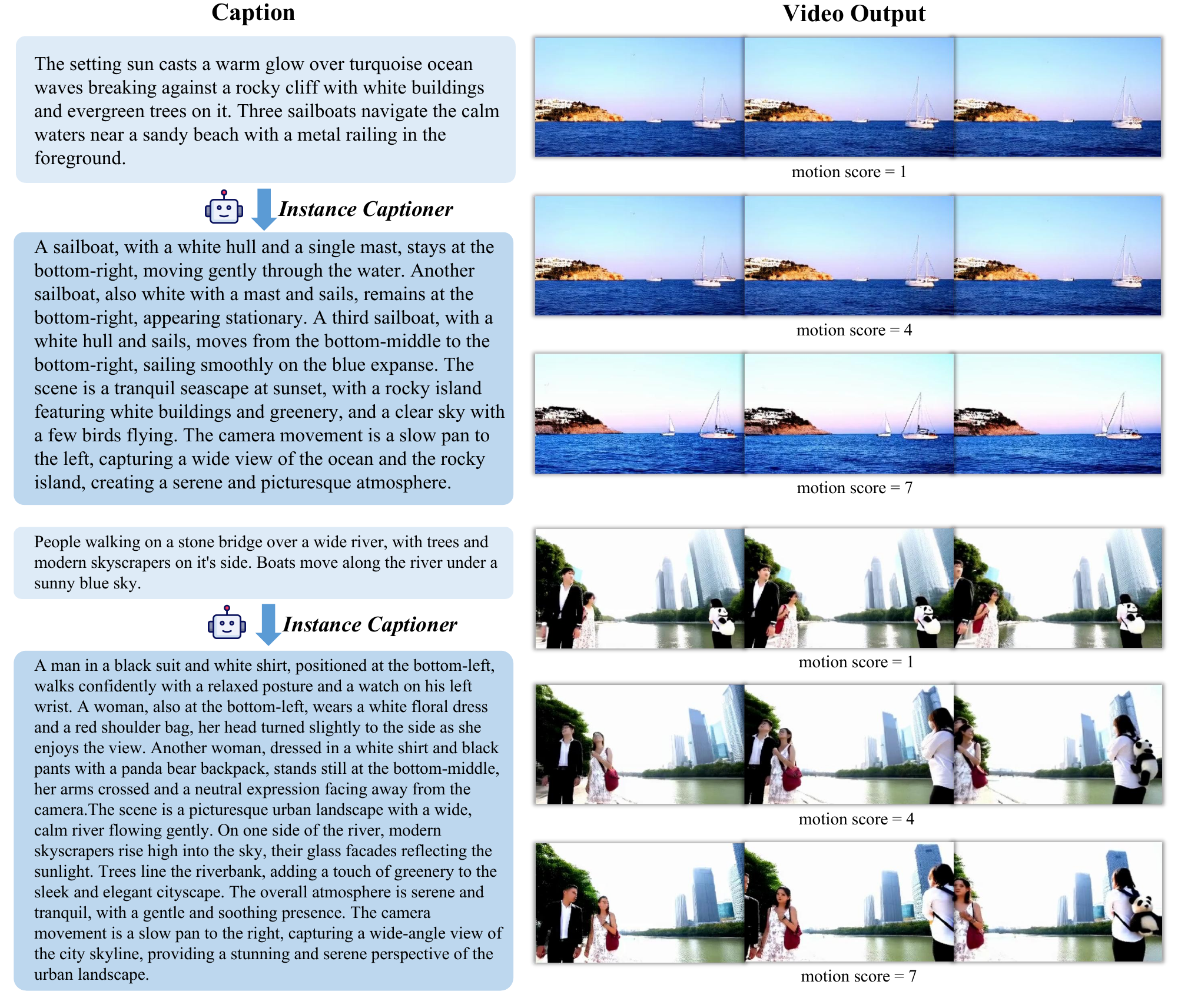}
    \caption{
    % instancecap的生成结果，我们展示了原始的caption和经过instance captioner处理后的caption，以及对于open-sora设置不同的motion score得到的结果，选择了第一帧中间帧和末尾帧进行展示。
    Generated results of InstanceCap, showing the original captions and the captions processed by the instance captioner. We also present the outputs from Open-Sora under different motion score settings, visualizing the first, middle, and last frames.
    }
    \label{fig:instancecap}
\end{figure*}

% instancecap是第一个提出instance-level视频生成概念的工作，但是并不是从模型层面，而是提出了一个instance-level的captioner和instance-level的训练数据，然后在开源的视频生成模型上进行微调。由于instancecap没有开源微调后的模型权重，因此我们这里比较的结果选择的时instancecap原文中描述的instancecaptioner和原始opensora相结合的输出视频。我们先用instancecaptioner对一段caption进行处理，然后直接输入到opensora中进行视频生成。

\noindent\textbf{InstanceCap} 
InstanceCap~\cite{fan2025instancecap} is the first work to introduce the concept of instance-level video generation, \textit{but it does not address this problem at the model-architecture level}. Instead, it proposes an instance-level captioner and constructs instance-level training data, which are then used to finetune existing open-source video generation models, like Open-Sora~\cite{opensora}. Since the finetuned model weights of InstanceCap are not publicly available, our comparison is conducted using the combination described in the original paper, namely the InstanceCap captioner with the original Open-Sora model. Concretely, we first process a given caption using the instance-level captioner, and then directly feed the resulting instance-aware caption into Open-Sora for video generation.

% 我们使用正文Figure.3中相同的测试用例按照上述方法对InstanceCap的性能进行了测试。由于GPU的影响限制，我们只生成了256px版本的视频。Figure.3左侧展示了InstanceCap的输出，InstanceCap接受一个文本和视频的输入然后输出instance-level video caption，这里为了节省空间我们没有展示原视频，可以看到instance-captioner的输出非常的细致，可以准确的描述每个instance的外表属性。但是由于文本描述固有的局限性，instance-captioner的输出仍然较难准确描述，或者说，较难辅助text encoder去准确理解不同instance之间的位置关系，交互关系或者更加复杂的关系。同时由于instance captioner本质是一个MLLM，将instance captioner所理解到的text-visual对应关系如果再次转换为文本，然后再经过video diffusion的text encoder处理，容易造成信息的损失和误差的累积。由于open-sora模型支持指定motion score，所以我们也生成了三组不同motion score的视频进行对比，我们发现，当motion score较大时输出的动态范围确实更加明显。不过两个测试用例的三组视频都不同程度地出现了缺少instance的情况（虽然这有可能是受限于分辨率但是由于硬件因素我们无法排查这一因素）。
We evaluate the performance of InstanceCap on the same test cases used in Figure.3 of the main text. Due to GPU limitations, we generate only the 256-px version of each video. The left side of Figure.~\ref{fig:instancecap} shows the outputs of InstanceCap. Given a text prompt and an input video, InstanceCap produces instance-level video captions. For brevity, we omit the original video frames. As shown, the instance-captioner generates highly fine-grained descriptions and accurately captures the visual attributes of each instance.

However, due to the inherent limitations of textual representations, these captions still struggle to precisely encode, or to effectively guide the text encoder to understand, spatial relations, inter-instance interactions, and other more complex relationships. Moreover, because the instance captioner is essentially an MLLM, translating its internal text–visual alignment back into textual form and then feeding this text again into the video diffusion model’s text encoder may introduce information loss and cumulative errors.

Since the Open-Sora model allows specifying a motion score, we additionally generate three sets of videos under different motion-score settings. We observe that higher motion scores indeed lead to videos with noticeably larger motion ranges. Nevertheless, across both test cases, all three generated versions exhibit varying degrees of missing-instance issues. Although this might be partially attributed to the reduced spatial resolution, our hardware constraints prevent us from isolating this factor.

\subsection{Evaluation on More Metrics and Benchmarks}
% 由于正文部分的空间有限，我们在这里额外进行了基于几组awesome的metrics和benchmarks on video generation的实验，简单介绍不同的metrics和benchmarks，记录并分析实验结果。
Due to space limitations in the main text, we provide \textit{additional experiments} here based on several widely recognized metrics and benchmarks for video generation, briefly introducing the different metrics and benchmarks, and recording as well as analyzing the experimental results.

% video score
% 

\noindent\textbf{VideoScore}
VideoScore~\cite{he2024videoscore} is a learned, fine-grained evaluation metric designed to approximate human feedback on video generation quality. It is trained on the VideoFeedback dataset, which contains tens of thousands of human-annotated videos rated across multiple dimensions, including visual quality, temporal consistency, dynamic richness, text-to-video alignment, and factual consistency. The resulting metric can generate scalar feedback scores for each dimension, providing a multi-faceted, human-aligned signal that correlates strongly with human judgments. Unlike simple distributional metrics (such as FVD) or single-aspect scores, VideoScore enables more nuanced, aspect-specific evaluation of generative video models, making it particularly useful for optimizing models in a human-preference-aware manner.
% videoscore评测模型共有两个版本，五个不同维度的指标，具体的实验结果如table3所示。
The VideoScore evaluation framework consists of two model versions and measures performance across five distinct dimensions. The detailed experimental results are presented in Table.~\ref{tab:vs} and Table.~\ref{tab:vs1.1}.

% 虽然在不同的评测模型版本下，评测结果有略微差异，但是并不影响整体质量的判断。在五个指标下instancev均达到了最优，在visual quality，temporal consistency和dynamic degree这三个指标下的优越性更加明显。这是由于提出的IMCA模块通过instance-level的trajectory信息来间接实现运动性和相机参数控制，提升了动态范围。同时STAPE通过赋予局部instances prompts以全局语义来实现时序一致性。并且，通过提升native cross-attention module对局部instances的关注度来提升视觉质量。

\noindent\textbf{VBench-Series} 
To enable comprehensive and standardized for video generation, the VBench-series of benchmarks provides a progressively refined suite of evaluation tools tailored for modern video generation systems.
\textbf{(1)} \textit{VBench}~\cite{huang2024vbench} introduced the first large-scale, multi-dimensional evaluation protocol specifically designed for video diffusion and autoregressive generators. Instead of relying solely on traditional metrics such as FVD or IS, VBench organizes evaluation into multiple capability dimensions, including \textit{appearance quality}, \textit{motion consistency}, \textit{temporal coherence}, \textit{subject identity preservation}, and \textit{text–video alignment}. It further provides human-annotated references and model-predicted proxies, enabling both automated and human-aligned scoring. 
\textbf{(2)} \textit{VBench++}~\cite{huang2024vbench++} builds upon VBench by \textit{adding evaluations for I2V models}. More importantly, VBench++ introduces an assessment of g\textit{enerative models' trustworthiness}, specifically measuring: how well they generate content that is fair across different cultures and demographics, and how effectively they avoid producing harmful or offensive content.
and demographics, and how effectively they avoid producing harmful or offensive content.
\textbf{(3)} \textit{VBench2}~\cite{zheng2025vbench} further introduces evaluations of generative models' \textit{intrinsic faithfulness}, such as whether the generated videos adhere to physical laws, commonsense reasoning, anatomical correctness, and compositional integrity.

% vbench中的部分指标使用的是预设的固定的指标对生成模型进行评测的，旨在建立一个统一而确定性的评价标准。但是这并不适用于instance-level的评测，所以我们只对vbench中可以对customized video outputs进行评测的指标对InstanceV和state-of-the-arts模型进行评测。同时由于instanceV并不是专门生成包含人的是视频的，因此vbench++中的伦理和公平性相关的指标以及vbench2中关于人体的指标我们没有进行评测。最终除了main text中的结果之外，我们选择了六个指标，前五个为vbench中的质保最后一个是vbench2中的指标。

Some of the metrics in VBench rely on predefined prompts and fixed evaluation criteria designed to provide a unified and deterministic benchmark for video generation models. However, such metrics are not suitable for instance-level evaluation. Therefore, we only include VBench metrics that support evaluating customized video outputs when comparing InstanceV with state-of-the-art models. 
In addition, since InstanceV is not specifically designed to generate videos containing humans, we exclude the ethics and fairness–related metrics in VBench++ as well as the human–centric metrics in VBench2. Beyond the results reported in the main text, we further evaluate six metrics in total: \textit{five quality-related metrics from VBench and one additional metric from VBench2}.

% 实验结果在Table4中所示，除了dynamic degree之外，其余的指标instanceV均超过了现有的sota模型。在Dynamic degree评价指标下，HunyuanVideo评测结果最佳，不过也因此使得subject consistency和background consistency中表现一般。通过实验我们确实发现hunyuanvideo的动态幅度表现很大。在不同指标下instanceV的综合表现更优，动态幅度和时序一致性的协调以及美学和图片质量都更好。同时在multi-view consistency指标中，instancev明显超过了大部分的现有sota模型，cogvideox因为生成的视频中包含相机参数变化的痕迹较少，所以指标比较接近instanceV。

The experimental results are summarized in Table~\ref{tab:vbench}. Except for the \textit{Dynamic Degree} metric, InstanceV outperforms all existing state-of-the-art models across the remaining metrics. For Dynamic Degree, HunyuanVideo achieves the best score, though this comes at the cost of weaker performance in \textit{Subject Consistency} and \textit{Background Consistency}. Our observations confirm that HunyuanVideo indeed produces videos with larger motion amplitudes. 
Overall, InstanceV demonstrates a more balanced and superior performance across diverse metrics, achieving a better trade-off between motion magnitude and temporal coherence, while also delivering stronger aesthetics and visual quality. In addition, for the \textit{Multi-View Consistency} metric, InstanceV significantly surpasses most state-of-the-art models. CogVideoX shows comparable performance mainly because its generated videos contain fewer apparent camera-motion variations, making its multi-view consistency score closer to that of InstanceV.

\begin{table*}[t]
    \centering
    \resizebox{\linewidth}{!}{
    \begin{tabular}{c|c|ccccc|c}
    \toprule
    % \textbf{Methods} & \textbf{Resolution}    & \textbf{Subject Consistency} $\uparrow$ &\textbf{Background Consistency}$\uparrow$&\textbf{Dynamic Degree}$\uparrow$&\textbf{Aesthetic Quality}$\uparrow$&\textbf{Imaging Quality}$\uparrow$ \\
    \textbf{Methods} & \textbf{Resolution} &
\makecell{\textbf{Subject}\\\textbf{Consistency}} $\uparrow$ &
\makecell{\textbf{Background}\\\textbf{Consistency}} $\uparrow$ & \textbf{Dynamic Degree}$\uparrow$&\textbf{Aesthetic Quality}$\uparrow$&\textbf{Imaging Quality}$\uparrow$ & \makecell{\textbf{Multi-View}\\\textbf{Consistency}} $\uparrow$ \\
     \midrule    
    Wan2.1  &  480P  & 0.97697 &   0.97170 &   0.14000 &0.57583 &  0.73665 & 0.43970 \\
    Wan2.2   & 720P &0.95958 & 0.95592 &   0.20000 & 0.51785 & 0.71953 & 0.46638\\
    HunyuanVideo   & 720P &0.96079 & 0.96021 &\textbf{0.57333}  &  0.58501& 0.69946&0.51205\\
    CogVideoX    &480P &0.98174& 0.97467& 0.20667 & 0.57144& 0.68036 &0.61747\\
     SkyReelsV2  & 540P &0.96392 &   0.96650 & 0.34000   &  0.60456  &  0.69523& 0.41249 \\
\midrule
        \multirow{2}{*}{\textbf{InstanceV(ours)}} &480P& 0.97237 & \textbf{0.97779} & 0.46000 & \textbf{0.60777} & 0.73499 & 0.55084\\
    &720P&\textbf{0.98211} & 0.97250 & 0.26000 & 0.57318 & \textbf{0.73875} & \textbf{0.62819} \\
      \bottomrule   
    \end{tabular}
    }
    \caption{Quantitative comparison of our proposed InstanceV with state-of-the-art video diffusion models using selected applicable metrics of \textit{VBench}(first five) and \textit{VBench2}(last one).}
    \label{tab:vbench}
\end{table*}

% fvmd

\section{More Details on Model Implementation}
\label{sec:model}
% 具体符号和Section.~\ref保持一致。
The notation below is kept consistent with the main text.
\subsection{IMCA}
% 对于输入IMCA的visual tokens，我们将它们沿着temporal dimension进行折叠，如图所示。这样的好处是两点
% 
For the visual tokens fed into the IMCA module, we fold them along the temporal dimension, as illustrated in Figure.2 of the main text. This design brings two major benefits: 
\begin{itemize}
    \item 
    % 更清晰的实现，将visual token沿着temporal dimension折叠意味着instance prompt tokens也要沿着temporal dimension折叠，因此不用对instance-level grounding-information做时间上的划分， 方便了IMCA的实现。
    By folding the visual tokens along the temporal dimension, the instance prompt tokens are naturally folded as well. Consequently, the instance-level grounding information does not require temporal partitioning, making the IMCA design more concise and efficient.
    \item 
    % 沿着temporal dimension这地visual token可以很大限度减小attention map的大小，从而减少attention机制耗费的时间和对GPU hardware的硬性要求。
    By folding visual tokens along the temporal dimension, the size of the attention map can be significantly reduced, which in turn decreases the computational overhead of the attention mechanism and relaxes the GPU memory and hardware requirements.
\end{itemize}

% 同时，由于native self-attention模块主要负责空间和时间一致性的建模，所以即使分开对frame进行instance-level grounding information的注入也不会对空间和时间一致性造成很大的损害。
\noindent Moreover, since the native self-attention module primarily handles the modeling of spatial and temporal consistency, injecting instance-level grounding information separately for each frame does not significantly compromise the overall spatial–temporal coherence.

\subsection{Attention Masks}
% 为了解决小物体消失的潜在问题，除了SAUG，我们还尝试了使用一个Float-version attention mask来进行解决。
To address the potential issue of small objects being overlooked, in addition to SAUG, we also experiment with using a \textit{float-version attention mask}.
Specifically, when a visual token corresponds to multiple instance tokens, we prioritize assigning higher attention weights to tokens corresponding to smaller instances. To implement this, we first sort the instance prompt tokens by their corresponding instance areas in descending order, and then modify the attention mask in MIV-CA to a floating-point version as follows:
\[ M^t_{(i, j)} =
\begin{cases}
j\times \mathrm{MaskBase} & \text{if } [\, V^t_i, I^t_j \,]=1,\\
-\mathrm{inf} & \text{if } [\, V^t_i, I^t_j \,]=0,
\end{cases} \]
,
here $\mathrm{MaskBase}$ is a predefined positive constant, whose magnitude determines the scaling factor applied to the attention map. In practice, we find that values such as 0.1 and 0.01 are sufficient to achieve the desired effect.
% 不过$\mathrm{MaskBase}$可能对不同的DiT-based video diffusion models都不一样，这取决于不同模型的DiTBlock的不同输出和中间结果的统计量信息。
However, $\mathrm{MaskBase}$ may \textit{vary across different DiT-based video diffusion models}, depending on the specific outputs and intermediate statistical information of each model's DiT blocks.

\begin{figure}[t]
    \centering
    \includegraphics[width=\linewidth]{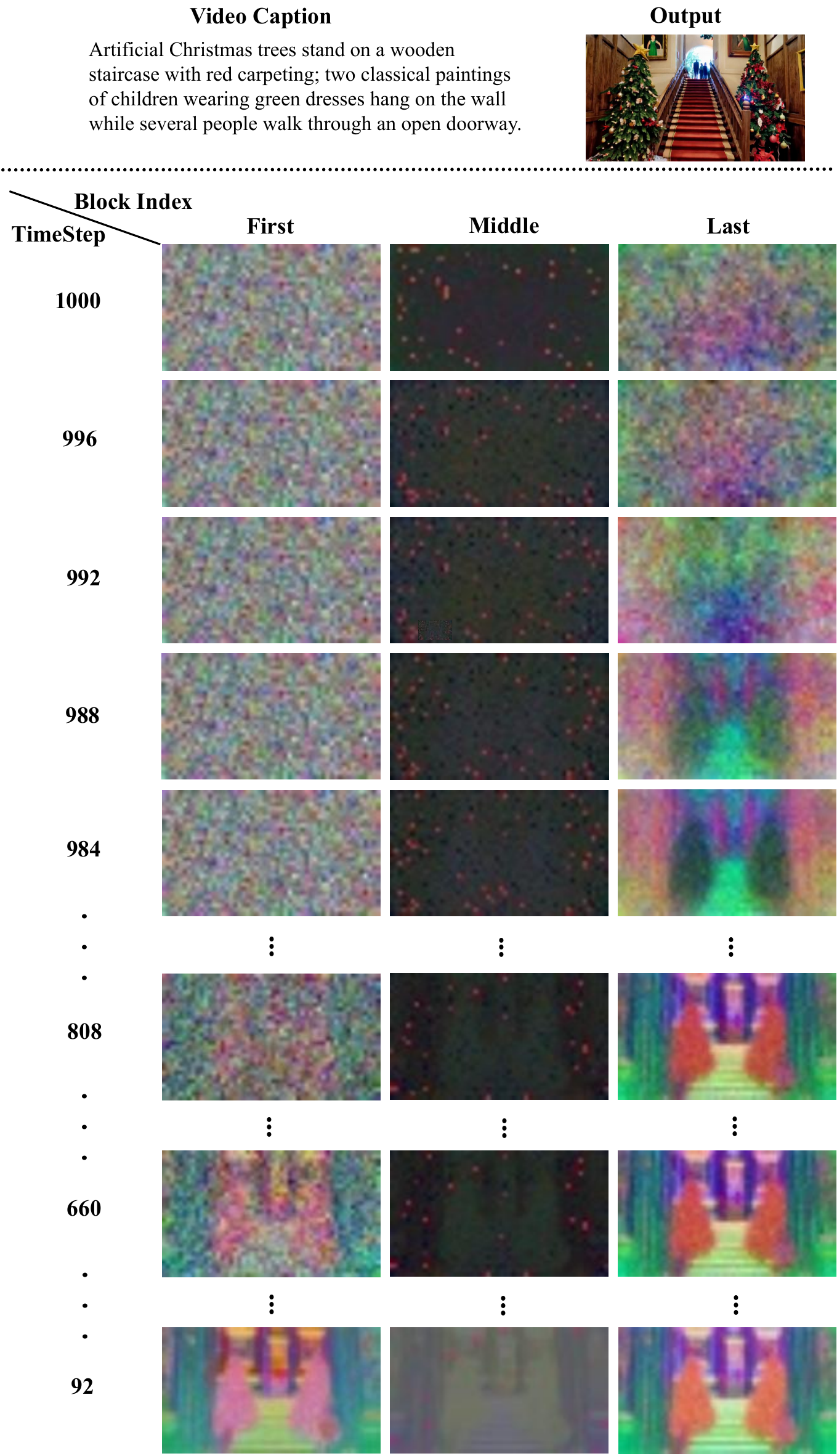}
    \caption{
    % 在不同的timestep下不同的ditblock输出visual tokens的PCA降维可视化，这里为了展示我们只取了第一个H*W的visual tokens。
    PCA visualization of the visual tokens produced by different backbone native DiT blocks at various timesteps. For illustration purposes, only the first $H\times W$ visual tokens are used.
    }
    \label{fig:ca}
\end{figure}

\subsection{STAPE}
% 正如FIgure2中所示，当使用flow-match scheduler时，不同的DiTBlock在不同的timestep完成的事情完全不同。我们发现，对于大部分的前向过程，timestep=984时，最后一个ditblock的输出已经初步具备比较清晰的语义，尽管细节仍然不完善。但是在接下的很多timestep内，除了最后一个ditblock，其余的ditblock的输出仍然不具备明确的语义，我们认为这是self-attention建模空间-时间一致性和cross-attention建立text-visual alignment对visual token造成不同作用的结果。所以，在不同的timestep对于不同的ditblock给予一个自适应的instance-level全局增强语义对于visual tokens内部的时间空间一致性和text-visual的alignment都有积极的效果。
As illustrated in Figure.~\ref{fig:cfg}, when using the flow-match scheduler, different DiT blocks perform entirely different functions across timesteps. We observe that for most of the forward process, at $\mathrm{timestep}=984$, the output of the last DiT block already exhibits reasonably clear semantic structure, although the fine details are still incomplete. However, over many subsequent timesteps, the outputs of the other DiT blocks, except for the final one, still lack explicit semantic meaning. We attribute this to the different influences exerted on visual tokens by self-attention, which models spatial–temporal consistency, and cross-attention, which establishes text–visual alignment.

Therefore, assigning an adaptive instance-level global semantic enhancement to different DiT blocks at different timesteps can positively impact both the spatial–temporal consistency within visual tokens and the alignment between text and visual representations.

\subsection{SAUG}
% 在实现SAUG时，我们比较了两种合理的实现。由于在unconditional forward时，需要留空instance prompts，如果只是简单的将对应的字符串留空，那么attention mask将不会对生成结果有任何的指导作用，因为在这种情况下所有的instance prompt tokens和padding tokens的编码都一样，那么投影后的value tokens都一样，所以进行attention后的结果也一样，无论attention mask是否参与计算。为了能够使得attention mask能够起到位置指导作用，需要为每个instance prompts提供一个没有明确的语义但是有一定区分度的编码方式，第一种简单的实现是扩充backbone的native text encoder的vocab和embedding，因为整个text encoder都需要freeze，所以这里只能使用类似xavier的初始化方式。第二种是由于text encoder在训练时由于不同predefined的任务而定义了类似 <句子开始> <句子结束>这种特殊的token，这类token在umT5中被命名为<extra_id_x>。这类token的数量通常在几十到一百之间，足够cover训练数据中单个视频可能存在的instances的数量了。复用<extra_id_x>可以避免扩充tokenizer的vocab和text encoder的embedding layer，并且在编码空间中，他们有着无意义但是特殊的语义，非常适合表示不同的instances。

When implementing SAUG, we compare \textit{two reasonable design choices}. During unconditional forward passes, instance prompts must be left empty. However, simply replacing instance prompts with empty strings is ineffective: the attention mask would provide \textit{no} positional guidance because all instance-prompt tokens and padding tokens collapse to identical embeddings. As a result, the projected value tokens become indistinguishable, and the output of the attention operation remains the same regardless of whether the attention mask is applied.

To ensure that the attention mask can meaningfully guide spatial positions, each instance prompt must be assigned an embedding that carries no explicit semantics but remains distinguishable across instances. We consider two practical solutions:
\textbf{(1)}	Extending the vocabulary of the native text encoder.
We can expand the backbone’s tokenizer and embedding matrix to include additional placeholder tokens. Since the entire text encoder is frozen, these embeddings must be randomly initialized (e.g., using Xavier initialization~\cite{glorot2010understanding}).
\textbf{(2)}	Reusing predefined special tokens in the text encoder.
Many pretrained text encoders define special tokens such as \texttt{<bos>}, \texttt{<eos>}, or the \texttt{<extra\_id\_x>} tokens in models like uMT5~\cite{xue2021mt5}. These tokens typically number in the dozens or even around one hundred, sufficient to cover the maximum number of instances in a single training video. Reusing \texttt{<extra\_id\_x>} avoids modifying the tokenizer or embedding layer, and these tokens naturally provide semantically meaningless yet mutually distinguishable embeddings, making them ideal for representing different instances.

% \section{Human Study}
% \label{sec:human}

\section{Potential Limitations and Future Study}
\label{sec:future}
\subsection{Instance-level Grounding Information}
% 一个可能的限制是，instanceV里支持的instance-level的grounding information形式较为单一，目前只支持bounding-box形式的spatial configuration和text-prompts形式的instance information；由于video diffusion model的训练和推理都较为computational hard，所以，我们主要针对模型结构进行了lightweight方向的设计，以使得模型能够 under limited computational budget起到更好的instance-level grounding效果。我们已经开始对这一扩展开展了具体的研究，由于正文篇幅有限，我们没有做出具体的论述。
% 在gligen和instancediffusion的基础上，我们还可以扩充更多的instance-level grouding information，比如对instance的属性，我们可以不只局限于文本描述，而是可以更具体地提供一张图片甚至是一段视频；对于spatial configuration，我们可以提供更加具体和专用的trajectory和motion信息。

One potential limitation of InstanceV lies in the relatively limited forms of instance-level grounding information it currently supports. At present, it only accommodates bounding-box-based spatial configurations and text-prompt-based instance attributes. Due to the computationally intensive nature of training and inference in video diffusion models, we primarily focused on designing a lightweight architecture to achieve effective instance-level grounding under a limited computational budget. We have already initiated preliminary investigations to extend this capability, but due to space constraints, these are not discussed in detail in the main text. 
Building upon frameworks such as GLIGEN~\cite{li2023gligen} and InstanceDiffusion~\cite{wang2024instancediffusion}, it is possible to further enrich instance-level grounding information. For example, instance attributes need not be restricted to textual descriptions; more detailed guidance could be provided via a reference image or even a short video. Similarly, for spatial configurations, more precise and dedicated trajectory and motion information could be incorporated.

\begin{figure}[tb]
    \centering
    \includegraphics[width=\linewidth]{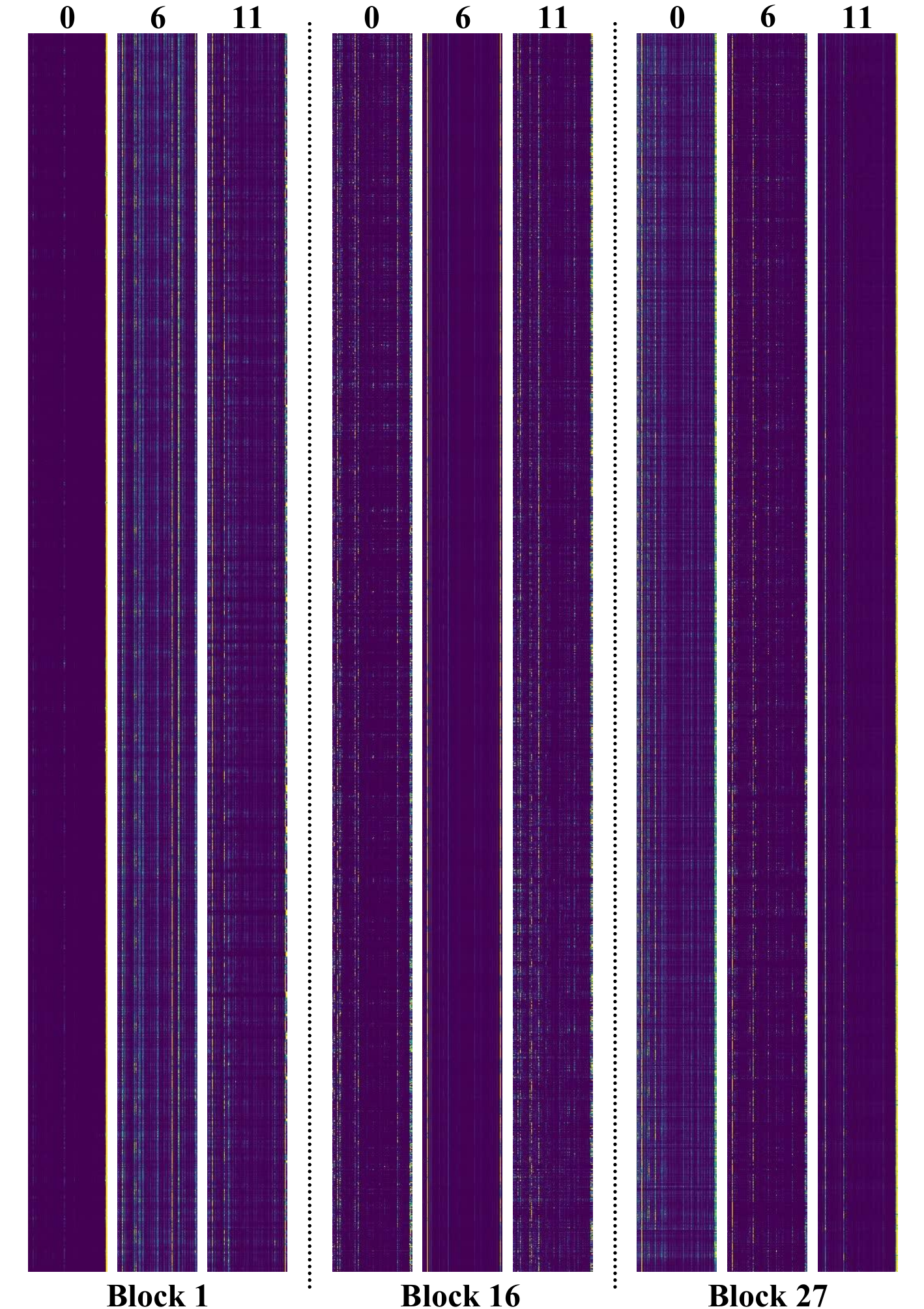}
    \caption{
    % $\mathrm{timestep}=984$时，不同ditblock中native cross-attention modules中不同attention heads对应的attention maps的可视化，我们去掉了padding tokens。
    Visualization of the attention maps from different attention heads in the native cross-attention modules across various DiT blocks at $\mathrm{timestep}=984$. Padding tokens are omitted.
     }
    \label{fig:ca_map}
\end{figure}

\begin{figure*}[t]
    \centering
    \includegraphics[width=\linewidth]{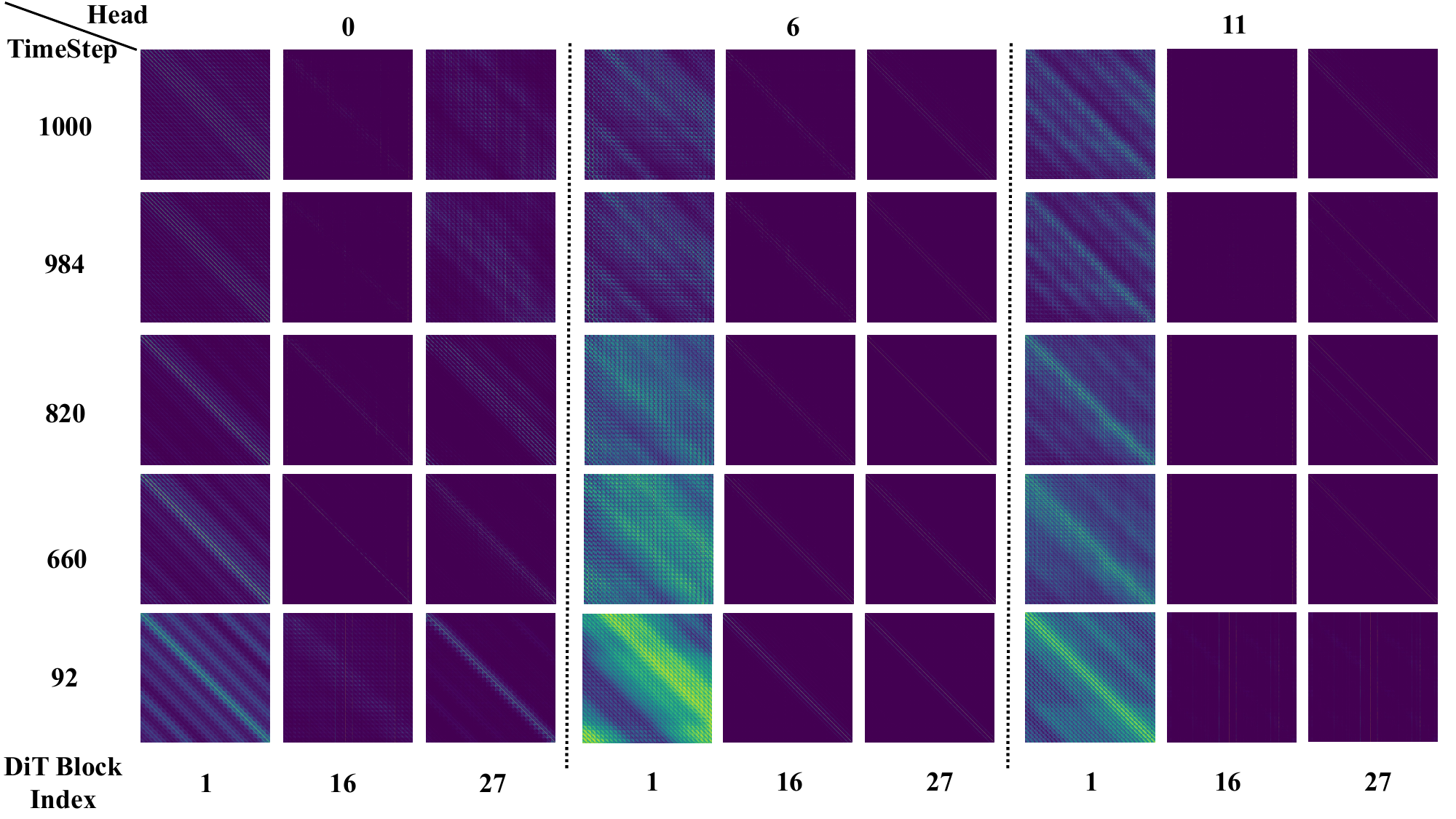}
    \caption{
    % native self-attention module中，在不同timestep，不同的ditblock的不同的attention head中attention map的可视化。
    Visualization of attention maps after normalization from different heads within various DiT blocks at different timesteps in the native self-attention module.
    }
    \label{fig:selfattn}
\end{figure*}

\subsection{Acceleration}
% 在InstanceV的基础上，我们甚至可以进一步对运算量做出优化。IMCA模块可能也不是每个DiTBlock都必须包含的，因为通过实验我们发现，在前几个DiTBlock层中，整个视频的layout就已经确定下来了，所以我们可以省略后面一些DiTBlocks中的IMCA。
% 进一步的，在多头注意力机制中，不同attention head的作用和贡献并不相同，我们可以进一步探索对每个ditblock中IMCA和cross-attention中的部分attention head做出取舍，以加速计算。
Building on InstanceV, we can further optimize computational efficiency. The IMCA module may not be required in every DiT block, as our experiments indicate that the overall video layout is largely established within the first few denoising timesteps. Therefore, IMCA can be omitted in later blocks without significant impact. 

% 正如figure2中所示，最后一个最后一个ditblock的输出在timestep=984后就基本上拥有稳定的语义了，尽管在接下来的一些timesteps内前面的ditblock的输出仍然没有非常清晰的语义信息。这引导我们向两方面思考：（1）我们是否可以对现有的video diffusion model在不同的video caption输出下的每个timesteps后的输出visual tokens做一个统计量分析，从而建立一个高效地routing策略，使得visual token，video caption和instance-level grounding information在最高效的timesteps和ditblocks中进行信息交互。（2）我们是否可以从native self-attention和native cross-attention相互作用的角度来在实例层面搭建一个直接高效的信息交互方式，尽可能减少二者对输出的冲突影响。
As shown in Figure.~\ref{fig:ca}, the output of the last DiT block stabilizes in terms of semantic content after $\mathrm{timestep}=984$, although the outputs of the preceding DiT blocks still lack clearly defined semantics for several subsequent timesteps. This observation motivates us to consider two directions: (1) whether we can perform a statistical analysis of the visual token outputs at each timestep under different video captions in existing video diffusion models, thereby establishing an efficient routing strategy that allows visual tokens, video captions, and instance-level grounding information to interact at the most effective timesteps and DiT blocks; (2) whether we can design a direct and efficient instance-level information exchange mechanism based on the interaction between native self-attention and native cross-attention, aiming to minimize potential conflicts between the two while preserving the quality of the outputs.

Furthermore, in the multi-head attention mechanism~\cite{vaswani2017attention}, different attention heads~\cite{gandelsman2023interpreting,lin2024mope} contribute unequally to the final output.
% 正如figure3所示，不同的ditblock中native self-attention modules在不同的time step的不同attention map的差异很大，进一步说明对最终输出的共享是不均衡的。有些attention map的语义很强，比如空间或者时间的语义，而有一些attention map只是简单的“自我关注”。
As shown in Figure.~\ref{fig:selfattn}, the native self-attention modules in different DiT blocks exhibit substantial variation in their attention maps across different time steps, indicating that sharing across the final output is inherently imbalanced. Some attention maps capture strong semantic information, such as spatial or temporal cues, while others primarily reflect simple “self-focused” patterns.
% 同样的，在native cross-attention modules中也会出现类似情况，正如Figure.~\ref{fig:ca}所示。由于空间有限我们只展示了$\mathrm{timestep}=984$的不同ditblocks中cross-attention module中不同head中的attention map，我们将padding tokens对应的attention map全部裁剪掉，只保留了有明确语义的部分。可以观察到与Figure.~\ref{fig:selfattn}中self-attention modules类似的现象，这是这里是cross-modal的情况。
A similar phenomenon also appears in the native cross-attention modules, as illustrated in Figure~\ref{fig:ca_map}. Due to space limitations, we only present the attention maps of different heads within the cross-attention modules across various DiT blocks at $\mathrm{timestep}=984$. We remove all attention maps corresponding to padding tokens and retain only the semantically meaningful regions. 

This suggests a potential direction to selectively prune or allocate IMCA and cross-attention heads in each DiT block, enabling further computational acceleration~\cite{xi2025sparse,yuan2024ditfastattn}.

\end{document}